\documentclass[10pt,journal,letterpaper,twocolumn,twoside]{IEEEtran} 

\usepackage{amsmath,amsfonts,bm,amssymb,psfrag,ifthen,color,subfigure}
\usepackage{mathrsfs}
\usepackage[justification=centering,font=footnotesize]{caption}
\usepackage{todonotes}

\allowdisplaybreaks[4]

\usepackage[dvips]{epsfig}
\usepackage[dvips]{graphicx}
\usepackage{amsmath,amsfonts,bm,amssymb,psfrag,ifthen,color,subfigure}
\usepackage{algpseudocode}
\usepackage{algorithm}
\usepackage{algorithmicx}
\usepackage{ifthen}
\usepackage{setspace}
\usepackage{cite}
\usepackage{url}
\usepackage{longtable}
\usepackage{stfloats}  
\usepackage{graphics,booktabs,color,subfigure}
\usepackage{float}
\usepackage{diagbox}
\usepackage{graphicx}
\usepackage{subcaption}

\usepackage[dvips]{epsfig}
\usepackage[dvips]{graphicx}
\usepackage{amsmath,amsfonts,bm,amssymb,psfrag,ifthen,color,subfigure,amsthm}
\usepackage{algpseudocode}
\usepackage{algorithm}
\usepackage{algorithmicx}
\usepackage{setspace}
\usepackage{cite}
\usepackage{url}
\usepackage{longtable}
\usepackage{stfloats}  
\usepackage{graphics,booktabs,color,subfigure}
\usepackage{float}
\usepackage{diagbox}
\usepackage{multirow}
\usepackage{makecell}
\allowdisplaybreaks[4]

	\usepackage{graphicx}
	
	\usepackage{makecell}
	\usepackage{graphicx}
	\usepackage{todonotes} 

	\usepackage{subfig}
	
\begin{document}	
				\title{Modeling and Performance Analysis for  Semantic Communications   Based on Empirical Results }

			
				\author{Shuai Ma,  Bin Shen, Chuanhui Zhang,  Youlong Wu,   Hang Li, Shiyin Li, Guangming Shi, and Naofal Al-Dhahir
				\thanks{Shuai Ma is with Peng Cheng Laboratory, Shenzhen 518066, China (e-mail:  mash01@pcl.ac.cn).}
				
%
%
%
%
			} 
					
			\maketitle
			\begin{abstract}

				Due to the black-box characteristics of deep learning based semantic encoders and decoders,	finding a tractable method for the performance analysis	of semantic communications  is a challenging problem.
				In this paper,  we propose an Alpha-Beta-Gamma (ABG) formula to  model the  relationship between the end-to-end   measurement and SNR, which can be applied for	both image reconstruction tasks and inference tasks.
				Specifically, for image   reconstruction tasks, the proposed ABG  formula  can well fit the commonly used DL networks,  such as  SCUNet,
				and Vision Transformer, for semantic encoding with the multi scale-structural similarity index measure (MS-SSIM) measurement. Furthermore, we find that the upper bound of the  MS-SSIM  depends
				on the number of quantized  output bits of   semantic encoders,  and we also propose a closed-form expression to fit the  relationship between the MS-SSIM and quantized output bits.
				To the best of our knowledge, this is the first theoretical   expression between  end-to-end performance metrics and   SNR  for  semantic communications.
				Based on the proposed ABG formula,  we investigate  an adaptive power control scheme  for semantic communications over random fading channels, which  can  effectively guarantee quality of   service (QoS) for semantic communications,  and then design the optimal power allocation scheme to maximize the  energy efficiency of the semantic communication system.
				Furthermore, by exploiting the bisection  algorithm,  we develop the power allocation scheme to maximize the minimum  QoS  of multiple users for OFDMA downlink semantic communication Extensive simulations verify the effectiveness and superiority of the proposed ABG formula and  power allocation schemes.

			\end{abstract}
			
			\IEEEpeerreviewmaketitle

			\section{Introduction}

			With the ever-increasing deployment of wireless intelligent applications networks, such as Digital Twins  \cite{Mihai_CST_2022} and Metaverse \cite{Lee_2024},  traditional communications are
			facing severe challenges of the unprecedented requirements on  spectrum and energy resources. To meet such requirements, semantic communications\cite{Dai_WC_2023},  a technique capable of significantly enhancing transmission efficiency, is regarded as a pivotal technology  , and has significant potentials in future sixth-generation (6G)
			networks\cite{Strinati_CN_2021}. Different from traditional communications that focuses on accurately transmitting all bits indiscriminately, semantic communication \cite{Lu_WC_2024}  focuses on extracting and conveying the  task-oriented meaning from the source data, which can dramatically reduce resource consumption of transmission  \cite{Lu_WC_2024,Patel_WE_2022,Lo_WCL_2023}.
			In \cite{Nam_ICASSP_2024}, a language-oriented semantic communication was designed by    integrating     in large language models (LLMs) and generative models, in which the source image are encoded into a text
			prompt via image-to-text (I2T) encoder, and the decoder progressively generates an image with received words by using Stable Diffusion.
			In \cite{Qiao_WCL_2024}, the authors developed a latency-aware semantic
			communications framework with pre-trained generative models, where the transmitter  extracts the semantic content of the	input signal in multiple semantic modalities, and then multi-stream scheme  are transmitted with  appropriate coding and communication
			techniques based on communication intent.
			Although the generative models based semantic communication schemes \cite{Nam_ICASSP_2024,Qiao_WCL_2024} greatly	improves communication efficiency, they can not be applied in key scenarios
			with semantic QoS performance requirements.

			Leveraging the advantages of machine learning, semantic communications has mainly adopted deep learning neural networks  for semantic coding, which demonstrates superior performance in point-to-point communication scenarios. In particular, the existing works on semantic communications can be categorized into two groups:  data reconstruction\cite{Yang_ICASSP_2023} and inference tasks \cite{Ma_TWC_2023}.

			Specifically, for data reconstruction tasks,
			various deep joint source-channel coding (JSCC) semantic communication
			schemes were designed for different data modalities, such as  text\cite{Xie_TSP_2021, Farsad_ACASSP_2018,Xie_JSAC_2020,Yi_TWC_2023,Jia_CL_2023}, speech\cite{Weng_JSAC_2021,Han_JSAC_2023}, image\cite{Huang_JSAC_2023,Zhang_JSAC_2023,Erdemir_JSAC_2023,Huang_IoT_2024,Kang_JSAC_2023} and video\cite{Tung_JSAC_2022, Jiang_JSAC_2023,Wang_JSAC_2023}.
			For example,  a transformer based on  semantic
			coding was proposed in \cite{Xie_TSP_2021}  by using sentence similarity for text transmission
			to deal with channel noise and semantic
			distortion.
			To accurately recover  speech information  at the semantic
			level, a   squeeze-and-excitation (SE) network based  semantic communication system
			was designed  in  \cite{Weng_JSAC_2021} for   robust transmission over  low signal-to-noise (SNR) channels.
			For image reconstruction tasks,
			a reinforcement learning based adaptive semantic
			coding (RL-ASC) approach was proposed in \cite{Huang_JSAC_2023} to jointly address the semantic
			similarity and perceptual performance issues.
			By  leveraging the perceptual quality of
			deep generative models (DGMs), the authors in \cite{Erdemir_JSAC_2023} proposed
			inverse JSCC and generative JSCC  schemes  for wireless image transmission.
			By integrating
			the   nonlinear transform coding into
			deep JSCC, a class of high-efficiency
			deep JSCC methods was proposed in   \cite{Dai_JSAC_2022}   to closely adapt
			to the source distribution.
			By transmitting some keypoints to represent the motions, an incremental redundancy hybrid automatic repeat request
			(IR-HARQ) semantic video system was designed  in \cite{Jiang_JSAC_2023} for  semantic video conferencing over time varying channels.
			By exploiting a nonlinear transform and conditional
			coding architecture to adaptively extract semantic features across
			video frames,
			a deep JSCC method was developed in\cite{Wang_JSAC_2023}
			to achieve	end-to-end video transmission over wireless channels.
			
			For inference tasks,  semantic communications focus on only transmitting the  information with the specific meanings  according to the tasks \cite{lee_Access_2019, Huang_JSAC_2022,Hu_TWC_2023,Shao_JSAC_2022, Xie_JSAC_2023}, instead of purely accurate data recovery.
			For example, in downstream inference tasks,   a variational information bottleneck (VIB) \cite{Tishby_arXiv_2000}  based  semantic encoding approach was proposed in \cite{Shao_JSAC_2022} to reduce  the feature transmission latency in
			dynamic channel conditions. A robust VIB based digital    semantic communications
			was designed in \cite{Xie_JSAC_2023}
			that can achieve better inference performance than the baseline methods
			with low communication latency. A  masked vector quantized-variational autoencoder (VQ-VAE) was
			designed in \cite{Hu_2022} with the noise-related masking strategy  to  improve    robustness against semantic noise.

			Note that the measurements of the existing semantic communication performance\cite{Zhang_TWC_2023} are  end-to-end, such as the classification accuracy for inference task and multi-scale-structural similarity index (MS-SSIM) for the image reconstruction task.
			Although it is well known that the performance of semantic communication depends on   transmission SNR, the relationship between the  end-to-end measurements and  transmission SNR is still not well discussed. The main reason is that  DL based semantic encoders  are  equivalent to highly complex nonlinear functions, and it is hard to use theoretical expressions to describe them. Hence, the theoretical basis for   physical   resource allocation has not been studied either.
			
			The previously discussed studies primarily focus on leveraging the feature representation capabilities of DL networks to enhance end-to-end performance in data reconstruction and classification tasks. However, due to the black-box nature of DL, it is hard to  establish the relationship between the end-to-end semantic transmission performance and SNRs,  which   results in a lack of theoretical tools for the design and analysis of semantic communications.
											
			In this paper,  we establish  the relationship between the end-to-end measurement and  SNR of semantic communications  for   both data reconstruction tasks and inference tasks.
			Then, we show that the proposed model is used to design optimal  power allocation  schemes with  a single user  and multiple users over random fading channels. The main contributions of this paper can be summarized as follows:
			\begin{itemize}
				\item 
				We propose an Alpha-Beta-Gamma (ABG) formula to model the  relationship between the performance metric and SNR for both the reconstruction and the inference tasks. Particularly, we reveal that the upper bound of MS-SSIM depends on the number of the quantized output bits of semantic encoders, and propose an upper bound with closed-form expression to establish  the relationship between the  MS-SSIM and the   number of the quantized bits. To the best of our knowledge, this is the first 
				empirical model between the end-to-end performance metrics and   SNR  of  semantic communications.

				\item
				
				We propose two power control strategies based on the ABG formula. 
				To guarantee the real-time quality of service over time-varying fading channel, we propose an adaptive power control schemes.
				We define the energy efficiency of semantic communications by  the ratio between image reconstruction quality and the total transmission energy consumption. Furthermore, we investigate the optimal power allocation to maximize the energy efficiency of the semantic communication system with the minimum rate requirement, and transmitted power constraint. 
				To solve this non-convex problem, we relax it into a concave-linear fractional problem. 
				Then,	by using the Dinkelbach-type algorithm, the original problem can be solved by optimizing a sequence of convex problems which converges to the global solution.

				\item 
				We extend the strategy design for the multi-user OFDMA semantic communication network with image reconstruction tasks.
				Specifically, our objective is to maximize the minimum MS-SSIM among multiple users under a total power constraint.
				To this end,  we design	a bisection  algorithm  to achieve the  optimal power allocation for the multi-user OFDMA semantic communication network.  Simulation  results show that the power allocation scheme outperforms the traditional power allocation scheme, which can  significantly improve the worst MS-SSIM of the multi-user OFDMA semantic communication network.
				
			\end{itemize}
			
			The rest of this paper is organized as follows. Section II introduces a  typical digital semantic communication system.  Section III presents the ABG formula for the semantic communication system.  In Section IV, we propose the optimal power allocation for    semantic communication systems. In Section V, the experimental results and analysis are presented. Finally, Section VI concludes the paper.

			\textit{{Notations}}:
			The following mathematical notations and symbols are used throughout this paper. Bold letters (e.g., $\bf{s}$) denote matrices (or multi-dimensional tensors), and regular letters (e.g., $p$) denote scalar. $\mathbf{I}$ refers to the identity matrix. The set of real numbers is denoted by $\mathbb{R}$, while $\mathbb{R}^{n \times m}$ represents $n$-by-$m$ real matrices. Similarly, $\mathbb{C}$ stands for complex numbers, and $\mathbb{C}^{n \times m}$ refers to $n$-by-$m$ complex matrices.
			The notations used in this paper are summarized in Table \ref{table1}.

		\begin{table}[htbp]
			\caption{Summary of key notations}
			\label{table1}
			\centering
				\resizebox{\linewidth}{!}{\begin{tabular}{|c|c|}
				\hline
				\rule{0pt}{7pt}\textbf{Notations}  &    \multicolumn{1}{c|}{\textbf{Meanings}} \\ \hline
				\rule{0pt}{6.5pt}$\mathbf{u} {\in  \mathbb{R}{^{ 3 \times H \times W   }} }$ &  Input data of TX  \\ \hline
				\rule{0pt}{6.5pt}$\mathbf{s} {\in \mathbb{R}{^{ \frac{{H}}{{16}} \times \frac{{W}}{{16}} \times 512   }} }$ &  Encoded semantic feature of TX \\ \hline
				\rule{0pt}{6.5pt}$\mathbf{b} {\in \mathbb{R}{^{{n_{\rm{b}}}}} }$ &  Quantified semantic features of TX  \\ \hline
				\rule{0pt}{6.5pt}$ { \mathbf{x}  \in \mathbb{C}{^{{n_{\rm{b}}}} } }$ & { Modulated signal } \\ \hline
				\rule{0pt}{6.5pt}$ { \mathbf{h} \in \mathbb{R}{^{{n_{\rm{b}}}}} }$ & {Channel gain  }\\ \hline
				\rule{0pt}{6.5pt}${\mathbf{z}  \in \mathbb{C}{^{{n_{\rm{b}}}}} }$ &  {Additive white Gaussian noise } \\ \hline
				\rule{0pt}{6.5pt}${\mathbf{y}  \in \mathbb{C}{^{{n_{\rm{b}}}}} }$ & { Received signal}  \\ \hline
				\rule{0pt}{6.5pt}${ \mathbf{\hat s} \in \mathbb{R}{^{{\frac{{H}}{{16}} \times \frac{{W}}{{16}} \times 512}}} }$ &  { Dequantized data }\\ \hline
				\rule{0pt}{6.5pt}${ \mathbf{\hat u} \in \mathbb{R}{^{3 \times H \times W}} }$ &  {Reconstructed data }\\ \hline
				\rule{0pt}{6.5pt}${{{{n_{\rm{b}}}}} }$ &  {Number of quantization bits }\\ \hline
				\rule{0pt}{6.5pt}  ${\alpha _{en}}$  &The parameters of semantic encoder of TX \\ \hline
				
				\rule{0pt}{6.5pt}  ${\alpha _{{\rm{de}}}}$  &The parameters of semantic decoder of TX \\ \hline
				
				\rule{0pt}{6.5pt}  ${{\cal F}_{\rm{en}}}\left(  \cdot  \right)$  & Semantic encoder  of TX   \\ \hline
				\rule{0pt}{6.5pt}  ${{\cal F}_{{\rm{de}}}}\left(  \cdot  \right)$ & Semantic decoder of RX  \\ \hline
				\rule{0pt}{6.5pt} $ Q\left(  \cdot  \right)$  &  Quantizer \\ \hline
				\rule{0pt}{6.5pt} {$ Q^{-1}\left(  \cdot  \right)$ } & Dequantizer  \\ \hline
				\rule{0pt}{6.5pt} $\Theta\left(  \cdot  \right)$   &  Modulator\\ \hline
				\rule{0pt}{6.5pt} ${\alpha(n_{\rm{b}})}$   &  {{Upper bound of end-to-end performance  in reconstruction tasks }}  \\ \hline
				\rule{0pt}{6.5pt} ${\alpha, \beta, \gamma, \tau}$   &  {{Fitting parameters of the ABG formula in reconstruction tasks}} \\ \hline
				\rule{0pt}{6.5pt} ${c_1, c_2, c_3, c_4}$   &  {{Fitting parameters of the ${\alpha(n_{\rm{b}})}$ in reconstruction tasks }} \\ \hline
				\rule{0pt}{6.5pt} ${\alpha_2(n_{\rm{b}})}$   & {{Upper bound of end-to-end performance  in inference tasks }}  \\ \hline
				\rule{0pt}{6.5pt} ${\alpha_2, \beta_2, \gamma_2, \tau_2}$   &  {{Fitting parameters of the ABG formula in inference tasks	}}  \\ \hline
				\rule{0pt}{6.5pt} ${c_5, c_6, c_7, c_8}$   &  {{Fitting parameters of the ${\alpha_2(n_{\rm{b}})}$ in inference tasks }} \\ \hline
				\rule{0pt}{6.5pt} $ {p_{\rm{cir}}} $  &  {The circuit power consumption}  \\ \hline
				\rule{0pt}{6.5pt} ${\psi _{{\rm{EE}}}}$ & {Energy efficiency of semantic communication system}  \\ \hline
			\end{tabular}}
		\end{table}

\begin{figure*}[!t]
	\centering
	\includegraphics[width=17cm]{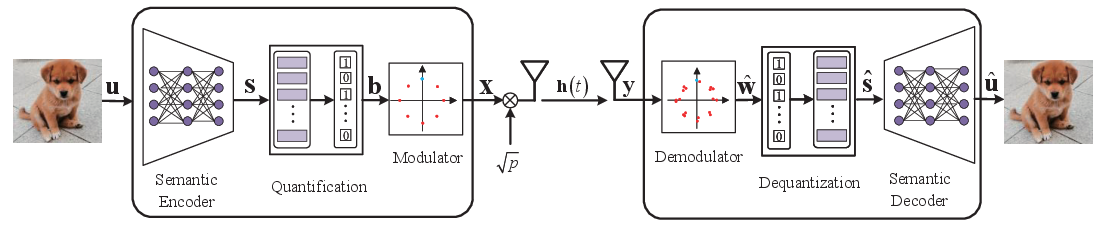}
	\caption{Typical  semantic   communication  system.}
	\captionsetup{justification=justified}
	\label{semantic_communication_system}
\end{figure*}

\section{Semantic Communication  System  Model}

Consider a typical point-to-point semantic   communication  system, as illustrated in Fig.\ref{semantic_communication_system}, where a semantic transmitter sends semantic information ${{\mathbf{s}}}$ of the input data ${{\mathbf{u}}} $ to a  receiver for inference   or data reconstruction tasks.
Specifically, via a semantic encoder ${\cal F}_{\rm{en}}\left(  \cdot  \right)$,  the input data ${{\mathbf{u}}}$ is encoded to the semantic feature ${{\mathbf{s}}}  $ as follows:
\begin{align}
{\bf{s}} = {{\cal F}_{\rm{en}}}\left( {{\bf{u}}} \right).
\end{align}

Generally, the  encoded  semantic feature    $\mathbf{s}$  is continuous, and the direct transmission of continuous feature representation  needs to be modulated with
analog modulation or a full-resolution constellation, which brings huge burdens for resource-constrained transmitter and poses implementation challenges on the current radio frequency (RF) systems. 
To be compatible with  digital transmission,   the semantic feature vector $\mathbf{s}$ is  quantized to a   ${{n_{\rm{b}}}}$-dimension binary     vector   ${\bf{b}} \in \mathbb{R}{^{{n_{\rm{b}}}}}$   as 
\begin{align}
	{\bf{b}} = Q(\mathbf{s}),
\end{align}
where $Q(\cdot)$  represents the quantization process, where by applying the one-bit quantization module \cite{Jiang_TC_2022}, the semantic feature $s$ is first transformed into a $n_{\rm b}$-dimensional vector using a fully connected (FC) layer and then quantized into a binary vector.

Furthermore,   the discrete vector $\bf{b}$ is modulated to  ${\bf{x}}$ as
\begin{align}
	\mathbf {\bf{x}} = \Theta \big(\bf{b}\big).
\end{align}

Then, ${\bf{x}}$ is transmitted to the receiver over random fading channel, and  the received signal is given by
\begin{align}
{\bf{y}} = {\bf{h}}\odot \sqrt {p} {\bf{x}} + {\bf{z}},
\end{align}
where $\odot $ represents the Hadamard product, $\bf{h}$ is channel gain, $ {p}$ denotes transmit power and  ${\bf{z}}\sim\mathcal{CN}\left(0,\sigma^2 \bf{I}\right)$ denotes the received  additive white Gaussian noise. $\sigma^2$ and $\bf{I}$ indicate the noise variance and the identity matrix, respectively.

Then, by applying the demodulator, the received signal $\mathbf{y}$ is demodulated to   $\mathbf {\hat{w}}$, i.e.,
\begin{align}
{\bf{\hat w}} = {\Theta ^{ - 1}}\left( {\bf{y}} \right).
\end{align}

Moreover, the dequantization process ${Q ^{ - 1}}(\cdot)$ is performed using a fully connected (FC) layer to restore the dimensionality of $ \bf{\hat{w}} $ to be consistent with that of $ \bf{s} $, resulting in the dequantized data $\bf{\hat{s}} $, i.e.,
\begin{align}
	{\bf{\hat s}} = {Q^{ - 1}}\left( {{\bf{\hat w}}} \right).
\end{align}

Furthermore, by applying semantic decoder ${{\cal F}_{{\rm{de}}}}\left(  \cdot  \right)$,   the receiver  decodes the  dequantized data $\mathbf {\hat{s}}$,   and  obtains  the reconstructed data   ${\hat {\bf{u}}}$ as 
\begin{align}\label{999}
{\bf{\hat u}} = {{\cal F}_{{\rm{de}}}}\left( { {\bf{\hat s}}} \right).
\end{align}

Due to the DL based encoders and decoders treating the process of extracting and recovering semantic information as a black-box, it cannot be precisely expressed mathematically. 
Note that, in the existing literature, the relationship between end-to-end performance metrics and transmission SNR has not been established, whether for inference tasks or data reconstruction tasks.
Thus, the theoretical basis of resource allocation for semantic communications is unknown, which leads to performance degradation in random fading channels.
To address this challenge,  we will investigate the relationship between  end-to-end performance measurements and  SNR, and propose a power alloation scheme for semantic communication systems in the following sections.

\section{ABG Modeling for Semantic Communications}

Since the deep  learning networks of the  semantic encoder and decoder  are generally  complex and  highly nonlinear,  it is  challenging to derive the theoretical relationship between  end-to-end performance measurements and  SNR.
However, based on a large number of experiments, we found that the end-to-end performance of semantic communication depends on both the  SNR and the number of quantization bits for both inference tasks and  image reconstruction tasks.

\subsection{Proposed ABG Modeling}
In  the semantic communication system,  MS-SSIM is one of the typical  end-to-end performance measurements for the image reconstruction task. After extensive experiments, we find that MS-SSIM and the transmission SNR satisfy a specific law: \emph{as SNR increases, the image reconstruction quality MS-SSIM increases rapidly at first, and then slowly increases until it reaches the upper bound.}
Therefore, the  relationship between  MS-SSIM and SNR  can be  fitted by the following ABG (Alpha-Beta-Gamma) formula:
\begin{align}
	{\rm{\varphi }}\left( {\rho ,{n_b}} \right) = \alpha \left( {{n_b}} \right) - \frac{\gamma }{{1 + {{\left( {\beta \rho } \right)}^\tau }}}, \label{ABG_MSSSIM}
\end{align}
where ${\rm{\varphi }}\left( {\rho ,{n_b}} \right)$ denotes the image reconstruction metric (MS-SSIM);  $ \alpha \left( {{n_b}} \right)> 0$ denotes the  upper bound  of the image reconstruction  quality;    parameters  $\beta > 0$  and $\gamma > 0$ depend on the deep learning network; and  $\rho$ denotes the transmission SNR. 
Furthermore, the  high nonlinearity of deep learning networks also makes it challenging to theoretically derive the relationship between the parameters $ \alpha $, $\beta$,  $\gamma$, and $\tau$ in of  ABG formula.

However, for  practical applications,  the values of the parameters $\alpha$, $\beta$, $\gamma$  and $\tau$  can be obtained by  using nonlinear least squares to fit   the   deep learning network of semantic encoders and decoders.
	
Moreover, the physical meanings of the parameters $\alpha$, $\beta$,  $\gamma$, and $\tau$  are given as	
	\begin{enumerate}
		\item $\alpha$ is the upper bound of the      image reconstruction quality, whose value depends on the number of output bits  of the digital semantic encoder. In other words,  the  physical meaning of $\alpha$ is the optimal performance of image reconstruction quality for given the number of output bits  of  the   semantic encoder.
		
		\item $\beta$ and  $\gamma$ are     tuning parameters related to the transmission SNR $\rho$, and whose values reflect  the impact of the transmission SNR on the    image reconstruction quality of  the semantic communication system.
		
		\item $\tau$ is a parameter that determines the    nonlinearity degree of the semantic communication system, whose value reflects the coefficient of change of image reconstruction quality with SNR of   the semantic communication system.
		
	\end{enumerate}

Furthermore,  the   performance upper bound of the image reconstruction  quality  $\alpha \left( {{n_{\mathrm{b}}}} \right) $ depends on the number of  quantized output bits of     the semantic encoder $n_{\mathrm{b}}$.
Specifically, we formulate the upper bound of  ABG formula $\alpha \left( {{n_{\mathrm{b}}}} \right) $ as follows:
   \begin{align}
   	\alpha \left( {{n_{\mathrm{b}}}} \right) = {c_1} - \frac{{{c_3}}}{{1 + {{\left( {{c_2}{n_{\mathrm{b}}}} \right)}^{{c_4}} }}} \label{upper_bound_MSSSIM},
   \end{align}
where parameters  $c_1 > 0$, $c_2 > 0$, $c_3 > 0$,  and ${c_4} > 0$ depend on the semantic encoder network.

Due to the high nonlinearity of DL based semantic encoder and decoder networks, it is challenging to theoretically derive and prove the accuracy of the ABG formula for measuring the end-to-end performance of semantic communication. Nonetheless, we can provide a qualitative analysis of the ABG formula: the quality of semantic communication transmission depends on the amount of semantic information received from the source. 
As the SNR increases, the mutual information between $s$ and $y$, as well as between $s$ and $x$, satisfies the following inequality:
\begin{align}I\left( {{\rm{s;y}}} \right) \le I\left( {{\rm{s;x}}} \right) \le H\left( {\rm{s}} \right),
\end{align}
where the mutual information terms $I(s; y)$ and $I(s; x)$ describe the shared information between the source and the received signal $y$ and the source and the encoded signal $x$, respectively. Specifically, $I(s; y)$ represents the amount of information the semantic decoder receives from the source, and $I(s; x)$ represents the information extracted by the semantic encoder from the source signal. The term $H(s)$ refers to the total amount of semantic information available from the source.
Therefore, as the SNR increases and since the semantic encoder-decoder is imperfect, there exists an upper limit on the transmitted information. This limit is precisely the optimal performance achievable by the semantic encoder-decoder. This is consistent with the upper bound on the ABG formula.

Note that,  the proposed ABG formula  can also be applied  in semantic communication systems for text  transmission  with the Bilingual Evaluation Understudy (BLEU) measurement\cite{Peng_GCC_2022}; speech transmission with the Perceptual Evaluation of Speech Quality (PESQ) measurement \cite{Xiao_ICASSP_2023}; video transmission with the Peak Signal-to-Noise Ratio (PSNR) measurement \cite{Yang_JSTSP_2021}, and etc. 	
In summary, the inference accuracy and  data reconstruction performance of semantic communications   initially increase rapidly with SNR due to significant noise reduction and the transition from a noise-dominated to a signal-dominated regime.  
The details reason that the    performance law are list as follows:
	\begin{itemize}
	\item  \emph{Impact of Noise Reduction}:
	At low SNR, the   received signal is dominated by noise, which severely degrades the quality of the transmitted signal. The semantic encoder and decoder struggle to extract and reconstruct meaningful information from the noisy signal, leading to low performance. As SNR increases, the noise   decreases, and the signal becomes clearer. This allows the decoder to more effectively decode and reconstruct the semantic information, resulting in a rapid increase in both reasoning accuracy and reconstruction performance.
	Specifically, as SNR increases, the reduction in noise has a significant impact, allowing the system to recover more semantic information from the received signal, which leads to a rapid improvement in performance.
	
	\item \emph{Transition from Noise-Dominated to Signal-Dominated Transmission}:
	When SNR is low, noise heavily influences the transmission, and the semantic communication systems  performance is limited by the ability to handle this noise. As SNR increases, the semantic communications transitions from a noise-dominated regime to a signal-dominated regime. In this transition phase, even a small increase in SNR can lead to a substantial improvement in performance, as the system becomes better at extracting the underlying semantic information.
	Once the signal begins to dominate over noise, the semantic communication systems can utilize the transmitted information more efficiently, leading to significant performance gains. However, once the semantic communication  system reaches this stage, further increases in SNR provide diminishing returns.
	
	\item  \emph{Performance Saturation}:
	As SNR continues to increase, the noise becomes negligible, and the semantic communication system operates in an almost noise-free environment. The performance of the semantic communication system is primarily limited by the capabilities of the encoder, decoder, and underlying model architecture, rather than by the channel noise. As a result, the performance improvements slow down, leading to the observed plateau.
	When noise is no longer a limiting factor, the semantic communication systems performance reaches a saturation point where it is primarily constrained by the model's design and learning capacity. Further increases in SNR do not lead to substantial performance gains because the system has already achieved near-optimal performance given the model's architecture.
	
	\item   \emph{Nonlinear Performance Gains}:
	The initial rapid improvement can also be attributed to the nonlinear behavior of neural networks and other components used in semantic communication systems. These models are often highly sensitive to changes in input quality. As SNR increases and the input signal quality improves, these models can extract features and semantic information more effectively, resulting in a sharp increase in performance. However, as the signal quality becomes consistently high, this nonlinear improvement tapers off, leading to more gradual performance increases.
	Neural networks with nonlinear activations can exhibit rapid performance gains when the quality of the input improves, but as the input becomes sufficiently clean (high SNR), the rate of improvement slows down.
	
\end{itemize}

\subsection{Evaluating of the  ABG Modeling}

To verify the accuracy of the proposed  ABG formula for semantic communication systems, we adopt the  common image reconstruction-oriented semantic encoding   networks  such as CNN \cite{Xie_2023}, SCUNet \cite{Zhang_2023_Practical}, Vision Transformer \cite{Dosovitskiy_2020_arXiV} and Swin Transformer \cite{Yang_TCCN_2024} to evaluate the fitting performance of  ABG formula,  and the corresponding model architectures and parameters are listed in Table \ref{Model_architecture}.

\begin{table*}[htbp]
	\centering
	\caption{Model architectures and parameters of CNN, SCUNet, Vision Transformer and Swin Transformer based semantic communication systems}
	\resizebox{0.9\textwidth}{!}{%
	\begin{tabular}{|c|c|c|c|c|c|c|c|}
		\hline
		\textbf{Model} & \textbf{Phase} & \textbf{Structure} & \textbf{Parameters} & \textbf{\makecell{Activation \\ function}} & \textbf{Normalization}  & \textbf{Sampling} & \textbf{\makecell{Learning\\ rate}} \\
		\hline
		\multirow{2}{*}{CNN { \cite{Xie_2023}} } & Encode & 4 $\times$ Conv2d  & \makecell{Kernel\_size=4$\times$4 \\ Stride=2 \\ Padding=1} & ReLU & \makecell{Batch\\ Normalization}& $\bm{\downarrow} $16 & $1 \times 10^{\textsuperscript{-3}}$  \\
		\cline{2-7}
		& Decoder & 4 $\times$ ConvTranspose2d  & \makecell{Kernel\_size=4$\times$4 \\ Stride=2 \\ Padding=1} & ReLU & \makecell{Batch\\ Normalization} & $\bm{\uparrow}$16   &  \\
		\hline
		\multirow{3}{*}{SCUNet  { \cite{Zhang_2023_Practical}} }& \multirow{3}{*}{Encoder} & 4 $\times$ Conv2d  & \makecell{Kernel\_size=4$\times$4 \\ Stride=2 \\ Padding=1} & LeakyReLU & \makecell{Batch\\ Normalization} & $\bm{\downarrow} $16  & $1 \times 10^{\textsuperscript{-4}}$ \\
		\cline{3-3}\cline{4-4}
		& & 4 $\times$ Swin-Conv Block & \makecell{Kernel\_size=1$\times$1 \\ Stride=1 \\ Padding=0 \\ Window\_size=2 } & &\makecell{ Layer\\Normalization} & & \\
		\cline{2-7}
		& Decoder & 4 $\times$ Swin-Conv Block & \makecell{Kernel\_size=1$\times$1 \\ Stride=1 \\ Padding=0 \\ Window\_size=2 } & LeakyReLU & \makecell{Layer \\ Normalization}& $\bm{\uparrow}$16 &   \\
		\cline{3-4}
		& &  4 $\times$ ConvTranspose2d  & \makecell{Kernel\_size=4$\times$4 \\ Stride=2 \\ Padding=1} & &\makecell{Batch\\Normalization} & & \\
		\hline
		\multirow{2}{*}{\makecell{Vision \\ Transformer { \cite{Dosovitskiy_2020_arXiV}} } } & Encoder & Vision Transformer block & \makecell{Depths=24 \\Num\_heads=16\\ Patch\_size=16} & ReLU & \makecell{Layer\\ Normalization} & $\bm{\downarrow} $16  & $1 \times 10^{\textsuperscript{-4}}$ \\
		\cline{2-7}
		& Decoder & Vision Transformer block & \makecell{Depths=24 \\Num\_heads=16\\ Patch\_size=16} & ReLU & \makecell{Layer\\ Normalization}& $\bm{\uparrow}$16  &   \\
		\hline
		\multirow{2}{*}{\makecell{Swin \\ Transformer { \cite{Yang_TCCN_2024}}}} & Encoder &  Swin Transformer block & \makecell{ Depths=[2,4,4,4] \\ Num\_heads=[2,4,4,8] \\ Window\_size=2} & ReLU & \makecell{ Layer\\ Normalization} & $\bm{\downarrow}$16  & $1 \times 10^{\textsuperscript{-4}}$  \\
		\cline{2-7}
		& Decoder &  Swin Transformer block & \makecell{ Depths=[4,4,4,2] \\ Num\_heads=[8,4,4,2] \\ Window\_size=2} & ReLU & \makecell{Layer\\ Normalization} & $\bm{\uparrow}$16 &   \\
		\hline
	\end{tabular}}
	\label{Model_architecture}
\end{table*}

The model architectures and parameters of CNN, SCUNet, Vision Transformer and Swin Transformer based semantic communication systems with reconstruction tasks are listed in Table \ref{Model_architecture}. 
In CNN based semantic communication systems, the semantic encoder includes four convolutional blocks for down sampling, in which the key parameters of the convolutional block include a kernel size of $4 \times 4$, a stride of $2$, and a padding of $1$. 
Moreover, a batch normalization is used between two layers. The downsampling rate of the semantic encoder is  $16$. 
The architecture of the  semantic  decoder is symmetric with that of the semantic encoder. 
Additionally, ReLU is chosen as the activation function, while Adam is utilized as the optimizer with a learning rate of $1 \times 10^{-3}$.

In SCUNet based semantic communication systems,  both convolutional blocks and Swin-conv blocks are utilized as the network's encoder. 
Within the convolutional blocks, the paper employs four convolutional blocks with a kernel size $4\times4$, a stride of $2$, and a padding of $1$, with batch normalization applied between two layers. 
In the remaining four Swin-conv blocks, each block consists of a convolutional block with a kernel size of $1 \times 1$, a stride of $1$, a padding of $0$ and a Swin Transformer block with a window size of $2$, utilizing layer normalization between two layers. The downsampling rate of the semantic encoder is $16$, and the decoder parameters are symmetrical with the encoder parameters. LeakyReLU is employed as the activation function, and AdamW serves as the optimizer with a learning rate set to $1 \times 10^{-4}$.

In Vision Transformer (ViT) based semantic communication systems, the ViT-Large model is employed. Within the ViT block, the patch size is set to $16$, the number of heads to $16$, and the depth to $24$. The downsampling rate of the semantic encoder is $16$, and the decoder parameters are symmetrical with the encoder parameters. ReLU is used as the activation function  with layer normalization. Additionally, AdamW is utilized as the optimizer, with a learning rate set to $1 \times 10^{-4}$.

In Swin Transformer based semantic communication systems, both the encoder and decoder utilize four Swin Transformer blocks. Within these Swin Transformer blocks, the window size of each block is set to $2$, and the depths are respectively set to $[2, 4, 4, 4]$, while the number of heads is respectively set to $[2, 4, 4, 8]$. The  downsampling rate of the semantic encoder is $16$, and the decoder parameters are symmetrical with the encoder parameters. Layer normalization is consistently applied for normalization across the network, with ReLU selected as the activation function. Furthermore, Adam is chosen as the optimizer, with a learning rate set to $1 \times 10^{-4}$.

\begin{figure}
	\centering
	\begin{minipage}[t]{0.25\textwidth}
		\centering
		\includegraphics[width=\textwidth]{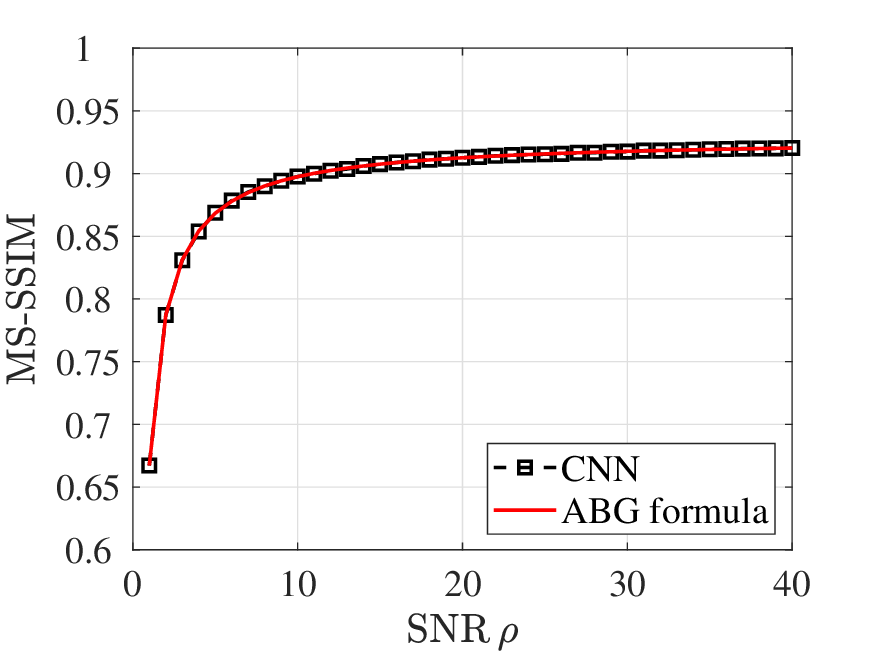} 
		\vskip-0.1cm
		\centering {(a)}
	\end{minipage}%
	\begin{minipage}[t]{0.25\textwidth}
		\centering
		\includegraphics[width=\textwidth]{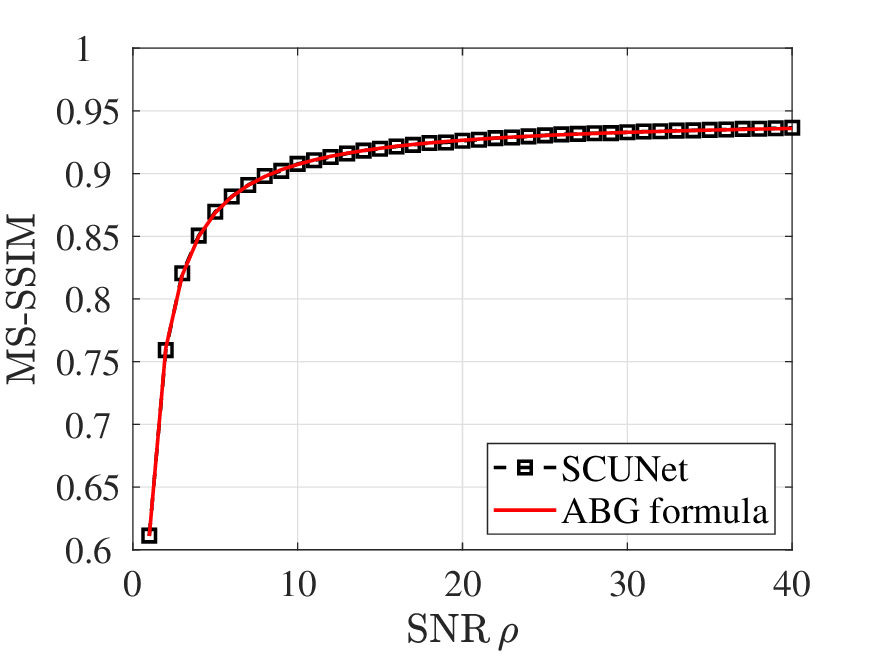} 
		\vskip-0.1cm
		\centering {(b)}
	\end{minipage}
	\begin{minipage}[t]{0.25\textwidth}
		\centering
		\includegraphics[width=\textwidth]{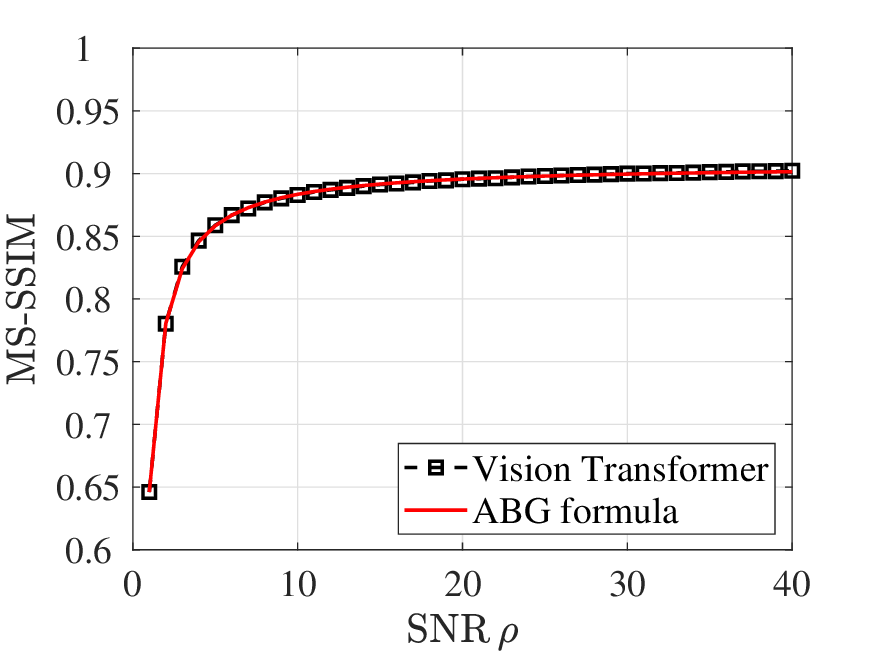} 
		\vskip-0.1cm
		\centering {(c)}
	\end{minipage}%
	\begin{minipage}[t]{0.25\textwidth}
		\centering
		\includegraphics[width=\textwidth]{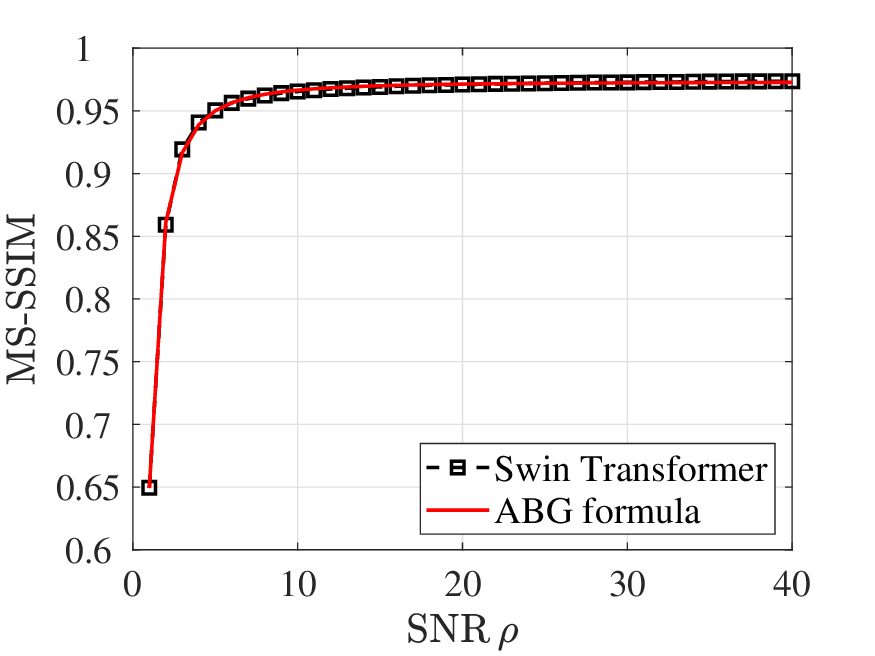} 
		\vskip-0.1cm
		\centering {(d)}
	\end{minipage}
	\captionsetup{justification=justified}
	\caption{ {MS-SSIM performance and the corresponding ABG formulas versus SNR $\rho$ in CIFAR-10 dataset  under Rayleigh channels: (a) CNN based semantic communication systems; (b)  SCUNet based semantic communication systems; (c)  Vision Transformer based semantic communication systems; (d)  Swin Transformer based semantic communication systems.}}
	\label{CIFAR10_SNR}
\end{figure}

Fig. \ref{CIFAR10_SNR} (a), (b), (c) and (d), demonstrate    the MS-SSIM performance of   CNN, SCUNet, Vision Transformer and Swin Transformer based semantic communication system versus SNR  in CIFAR-10 dataset under Rayleigh channels, respectively, where the dashed lines with squares are the  performance curves of the corresponding semantic encoding   networks, and the solid lines are the   curves  of  ABG formula.
As shown in Fig. \ref{CIFAR10_SNR} (a), (b), (c) and (d), the ABG formula   can well fit the  performance curves of CNN, SCUNet, Vision Transformer and Swin Transformer based  semantic communication systems.
Moreover,  with increasing SNR, the MS-SSIM first rises rapidly, and then approaches the performance upper bound $ \alpha \left( {{n_{\mathrm{b}}}} \right)$.

Furthermore, to quantitatively measure  the fitting errors  between the ABG formula  and the semantic communication systems  performance, we adopt the sum of squares error (SSE) to evaluate the goodness of model fitting\cite{Thinsungnoena_L_2015} as follows:
 \begin{align}\label{sse}
\varsigma  = {\sum\limits_{i = 1}^n {{g_i}\left( {{t_i} - {{\bar t}_i}} \right)} ^2},
\end{align}
where ${g_i}$ denotes the weight of the $i$-th data point, ${t_i}$ denotes the true value, and ${{\bar t}_i}$ stands for the fitted value.
Specifically, let $\varsigma_1$ denote SSE of the ABG formula.
The closer the value of  $\varsigma_1$  is to 0, the better the fitting performance of the ABG formula is.
Moreover,  the values of the ABG formula parameters $ \alpha \left( {{n_{\mathrm{b}}}} \right)$, $\gamma$, $\beta$,   $\tau_1$ and SSE $\varsigma_1$ are listed in Table \ref{snr_fit}.

\begin{table}[htbp]
	\captionsetup{justification=justified}
	\caption{{The parameters of ABG formula for fitting the CNN, SCUNet, Vision Transformer and Swin Transformer based  semantic communication systems with MS-SSIM in CIFAR-10 dataset under Rayleigh channels, and the corresponding SSE fitting errors. }}
	\centering
	\resizebox{\linewidth}{!}{%
		\begin{tabular}{|c|c|c|c|c|c|}
			\hline
			\rule{0pt}{8pt}Model &  $ \alpha \left( {{n_{\mathrm{b}}}} \right)$ & $\gamma$  &  $\beta$ & $\tau$ & Fitting errors $\varsigma_1$  \\ \hline
			\rule{0pt}{7.5pt} CNN &  {0.92} & {2.08} &  7.28 & {0.97} & {$2.56 \times 10^{\textsuperscript{-6}}$ }\\ \hline
			\rule{0pt}{7.5pt} SCUNet & {0.94} & {1.29}  & {2.70} & {1.06} &  {$9.22 \times 10^{\textsuperscript{-6}}$ }\\ \hline
			\rule{0pt}{7.5pt} Vision Transformer & {0.90} & {371.60} & 1055 & {1.04} & {$9.47 \times 10^{\textsuperscript{-6}}$ }\\ \hline
			\rule{0pt}{7.5pt} Swin Transformer  &  {0.97} & 1.36 & 1.91 & {1.79}  & {$ 3.04 \times 10^{\textsuperscript{-5}}$} \\ \hline
	\end{tabular}}
	\label{snr_fit}
\end{table}

Table \ref{snr_fit} shows the parameters of the  ABG   formulas for fitting the CNN, SCUNet, Vision Transformer and Swin Transformer based  semantic communication systems with MS-SSIM in CIFAR-10 dataset under Rayleigh channels, and the corresponding SSE fitting errors $\varsigma_1$.
As listed in Table \ref{fitting_error}, the SSE  fitting errors of  ABG   formulas for   the CNN, SCUNet, Vision Transformer and Swin Transformer based  semantic communication systems are  $2.5629 \times 10^{\textsuperscript{-6}}$, $9.2259 \times 10^{\textsuperscript{-6}}$, $9.4744 \times 10^{\textsuperscript{-6}}$ and $ 3.042 \times 10^{\textsuperscript{-5}}$ respectively, which   verify the accuracy of  the ABG formula.

%
%
%

\begin{figure}[!t]
	\centering
	\begin{minipage}[t]{0.25\textwidth}
		\centering
		\includegraphics[width=\textwidth]{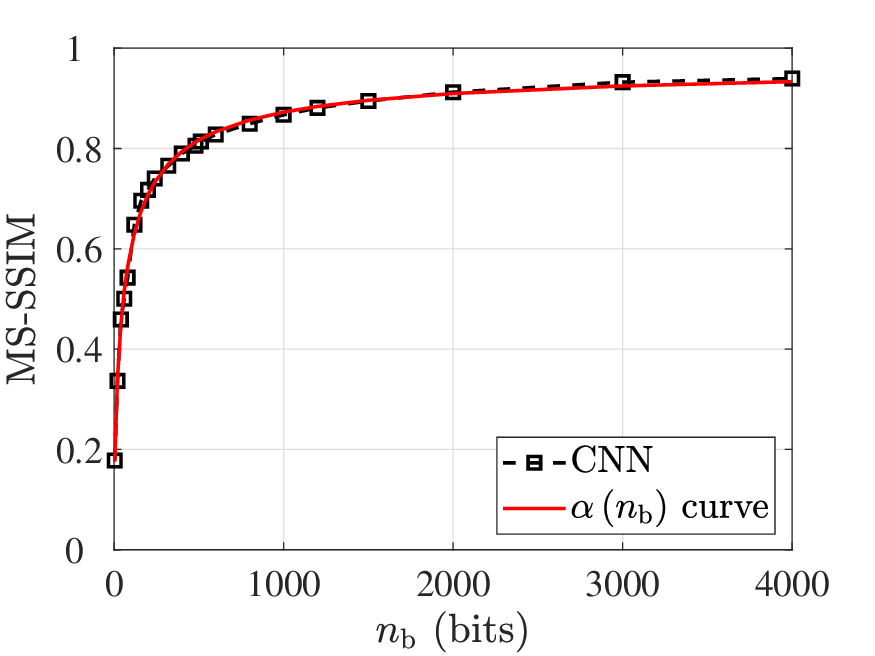}
		\vskip-0.1cm
		\centering {(a)}
	\end{minipage}%
	\begin{minipage}[t]{0.25\textwidth}
		\centering
		\includegraphics[width=\textwidth]{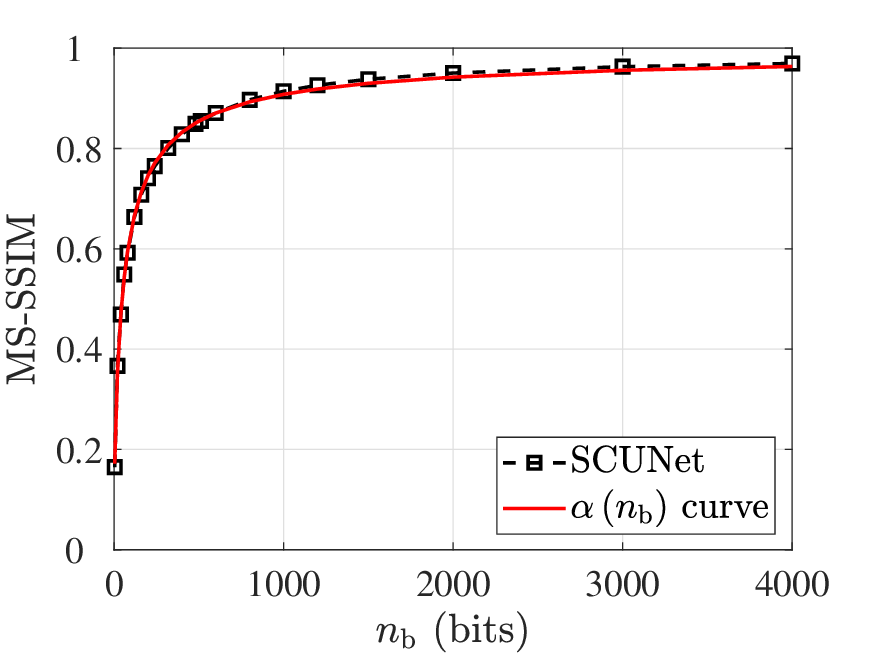}
		\vskip-0.1cm
		\centering {(b)}
	\end{minipage}
	
	\begin{minipage}[t]{0.25\textwidth}
		\centering
		\includegraphics[width=\textwidth]{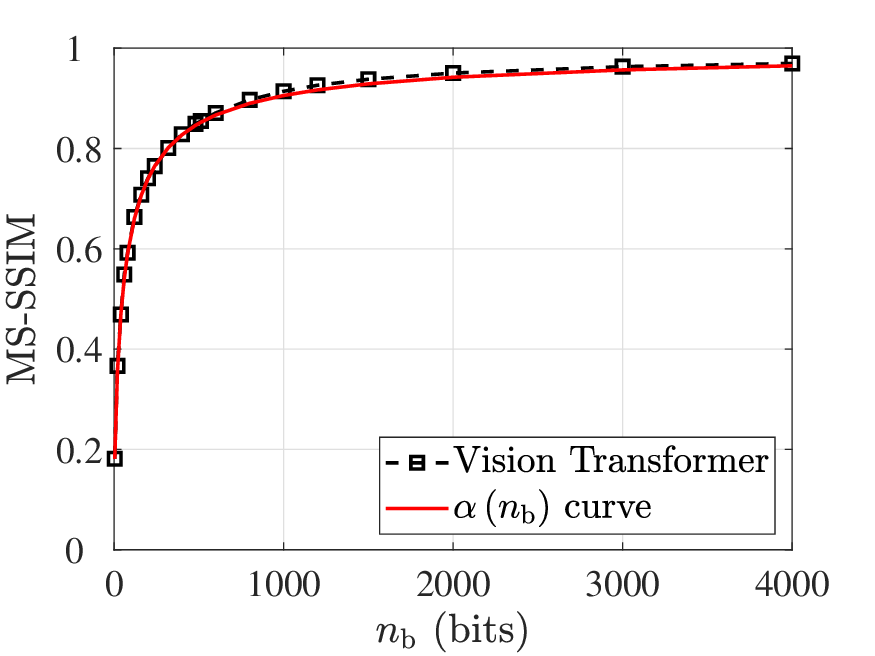}
		\vskip-0.1cm
		\centering {(c)}
	\end{minipage}%
	\begin{minipage}[t]{0.25\textwidth}
		\centering
		\includegraphics[width=\textwidth]{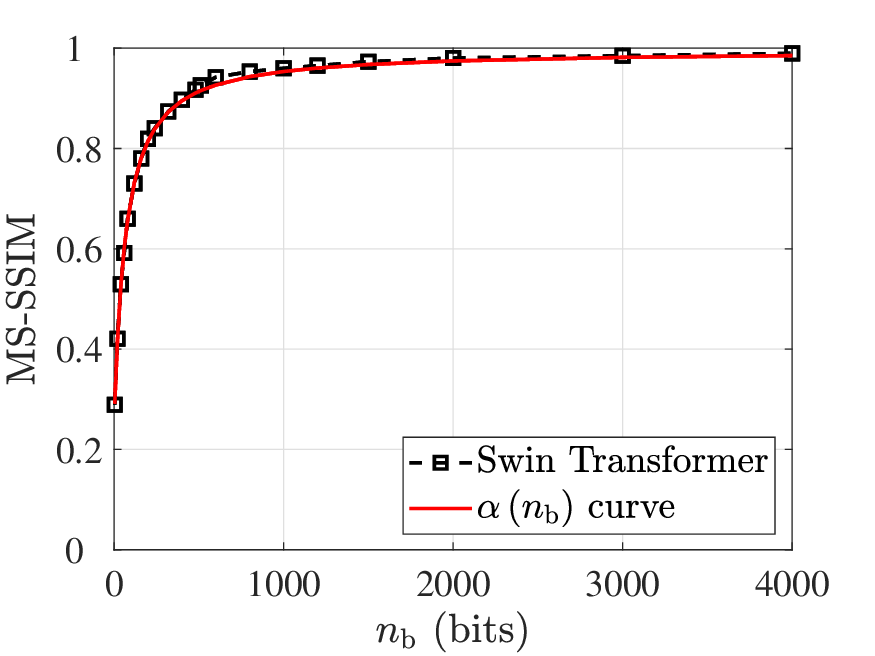}
		\vskip-0.1cm
		\centering {(d)}
	\end{minipage}
	\captionsetup{justification=justified}
	\caption{ MS-SSIM performance  versus the number of quantized bits $n_{\mathrm{b}}$ in CIFAR-10 dataset  under Rayleigh channels: (a) CNN based semantic communication systems; (b) SCUNet based semantic communication systems; (c)  Vision Transformer based semantic communication systems; (d) Swin Transformer based semantic communication systems.}
	\label{CIFAR10_BIT}
\end{figure}

Moreover,  we evaluate   the accuracy of parameter $\alpha \left( {{n_{\mathrm{b}}}} \right) $ in \eqref{upper_bound_MSSSIM} for fitting the  upper bound of MS-SSIM.
Fig. \ref{CIFAR10_BIT} (a), (b), (c) and (d), demonstrate the  upper bound of  MS-SSIM performance of   CNN, SCUNet, Vision Transformer and Swin Transformer based semantic communication system versus  the number of quantized bits $n_{\mathrm{b}}$ in CIFAR-10 dataset under Rayleigh channels respectively, where
the dashed lines with squares are the  performance curves of the corresponding semantic encoding   networks, and the solid lines are the   curves  of    $\alpha \left( {{n_{\mathrm{b}}}} \right) $ in \eqref{upper_bound_MSSSIM}.
As shown in Fig. \ref{CIFAR10_BIT}(a), (b), (c) and (d),   $\alpha \left( {{n_{\mathrm{b}}}} \right) $ in \eqref{upper_bound_MSSSIM}  can well fit the upper bound of the MS-SSIM of CNN, SCUNet, Vision Transformer and Swin Transformer based  semantic communication systems.

Table \ref{upbound_fitting_error} illustrates the parameters of  $\alpha \left( {{n_{\mathrm{b}}}} \right) $ for fitting the upper bound of the MS-SSIM of the CNN, SCUNet, Vision Transformer and Swin Transformer based  semantic communication systems in CIFAR-10 dataset under Rayleigh channels, and the corresponding SSE fitting errors $\varsigma_2$.
As shown in Fig. \ref{CIFAR10_BIT}(a), (b), (c) and (d),   $\alpha \left( {{n_{\mathrm{b}}}} \right) $ can well fit the upper bound of MS-SSIM of CNN, SCUNet, Vision Transformer and Swin Transformer based  semantic communication systems under Rayleigh channels, respectively,
As listed in Table \ref{upbound_fitting_error}, the SSE  fitting errors $\varsigma_2$ of the upper bound  $\alpha \left( {{n_{\mathrm{b}}}} \right) $ for   the CNN, SCUNet, Vision Transformer and Swin Transformer based  semantic communication systems are  $1.19911 \times 10^{\textsuperscript{-4}}$, $2.23744 \times 10^{\textsuperscript{-5}}$, $2.01312 \times 10^{\textsuperscript{-5}}$ and $1.00736 \times 10^{\textsuperscript{-5}}$, respectively, which verify the accuracy of   $\alpha \left( {{n_{\mathrm{b}}}} \right) $ in \eqref{upper_bound_MSSSIM}.

 \begin{table}[htbp]
	\captionsetup{justification=justified}
	\caption{ {The  parameters of $\alpha \left( {{n_{\mathrm{b}}}} \right) $     for fitting upper bounds of the MS-SSIM of the the CNN, SCUNet, Vision Transformer and Swin Transformer based  semantic communication systems in CIFAR-10 dataset under Rayleigh channels, and the corresponding SSE fitting errors.}}
	\centering
	\resizebox{\linewidth}{!}{%
		\begin{tabular}{|c|c|c|c|c|c|}
			\hline
			\rule{0pt}{8pt}Model &  $ c_1 $ &  $c_2$ &  $c_3$ & $c_4$ & fitting errors $\varsigma_2$  \\ \hline			
			\rule{0pt}{7.5pt} CNN & {0.96} & {0.89} & {0.01} & {0.75} &  {$1.19 \times 10^{\textsuperscript{-4}}$} \\ \hline
			\rule{0pt}{7.5pt} SCUNet & {0.99} & {0.92} & {0.01} & {0.79} &  {$4.26 \times 10^{\textsuperscript{-5}}$ }\\ \hline
			\rule{0pt}{7.5pt} Vision Transformer & {0.99} & {0.91} & {0.01} & {0.77} &  {$2.23 \times 10^{\textsuperscript{-5}}$ }\\ \hline
			\rule{0pt}{7.5pt} Swin Transformer  &  {0.99} & {0.74} & {0.01} & {1.04}  &
			{$1.00 \times 10^{\textsuperscript{-5}}$} \\ \hline
	\end{tabular}} 	\label{upbound_fitting_error}
\end{table}

\begin{figure}[t]
	\begin{minipage}{0.25\textwidth}
		\centering
		\centerline{\includegraphics[width=\textwidth]{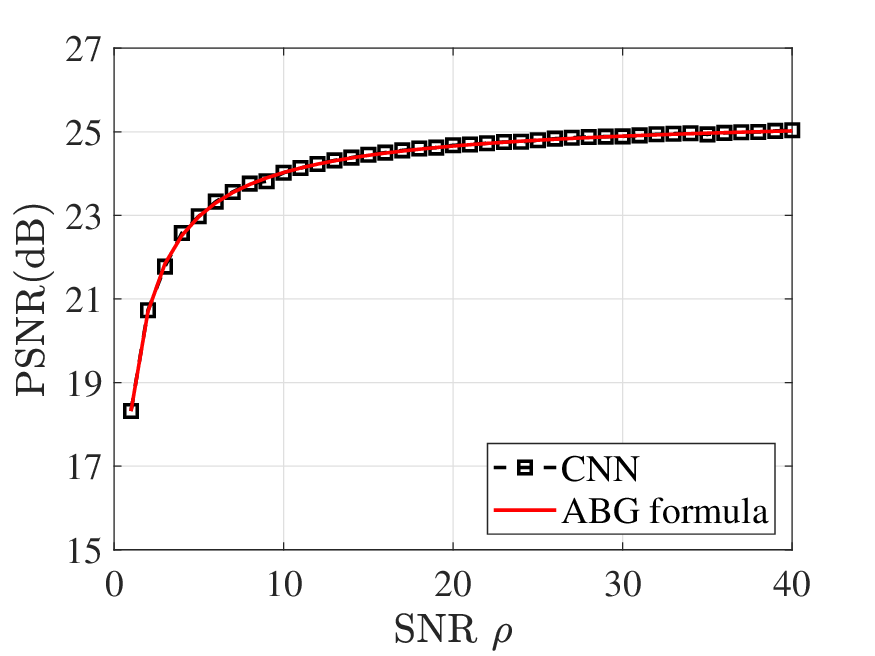}}
		\vskip-0.1cm
		\centerline{(a) }
	\end{minipage}%
	\begin{minipage}{0.25\textwidth}
		\centering
		\centerline{\includegraphics[width=\textwidth]{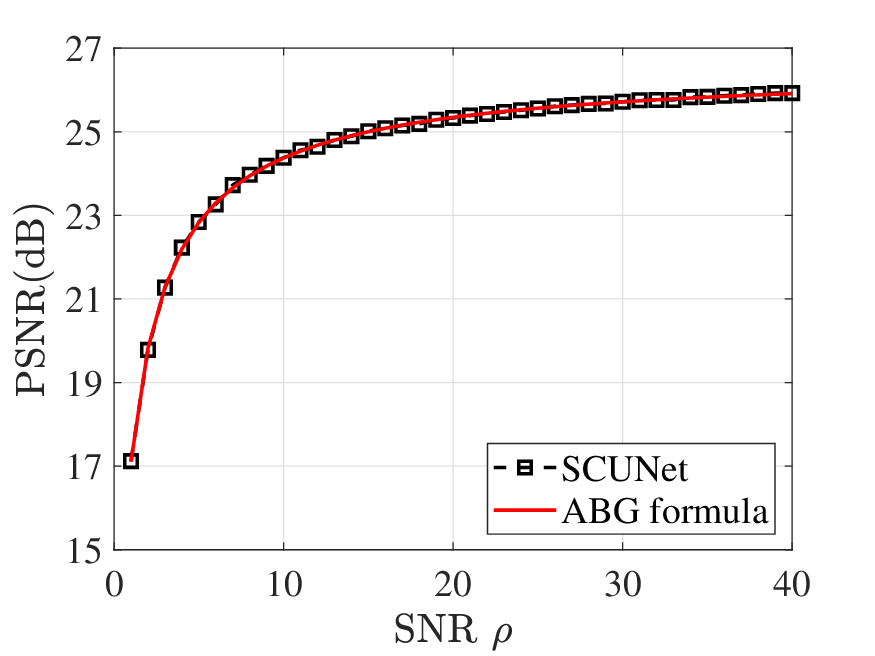}}
		\vskip-0.1cm
		\centerline{(b) }
	\end{minipage}

	\begin{minipage}{0.25\textwidth}
		\centering
		\centerline{\includegraphics[width=\textwidth]{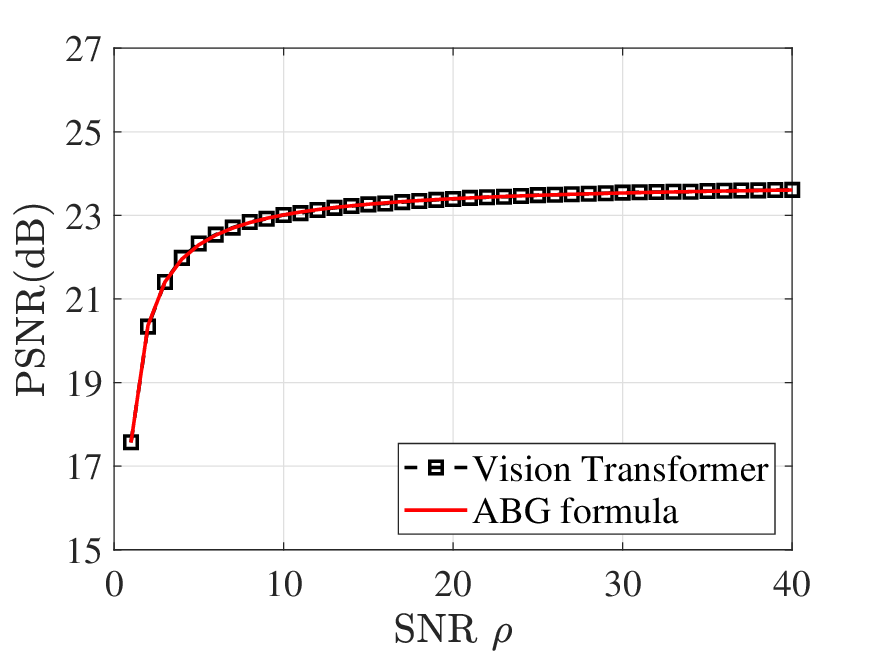}}
		\vskip-0.1cm
		\centerline{(c)}
	\end{minipage}%
	\begin{minipage}{0.25\textwidth}
		\centering
		\centerline{\includegraphics[width=\textwidth]{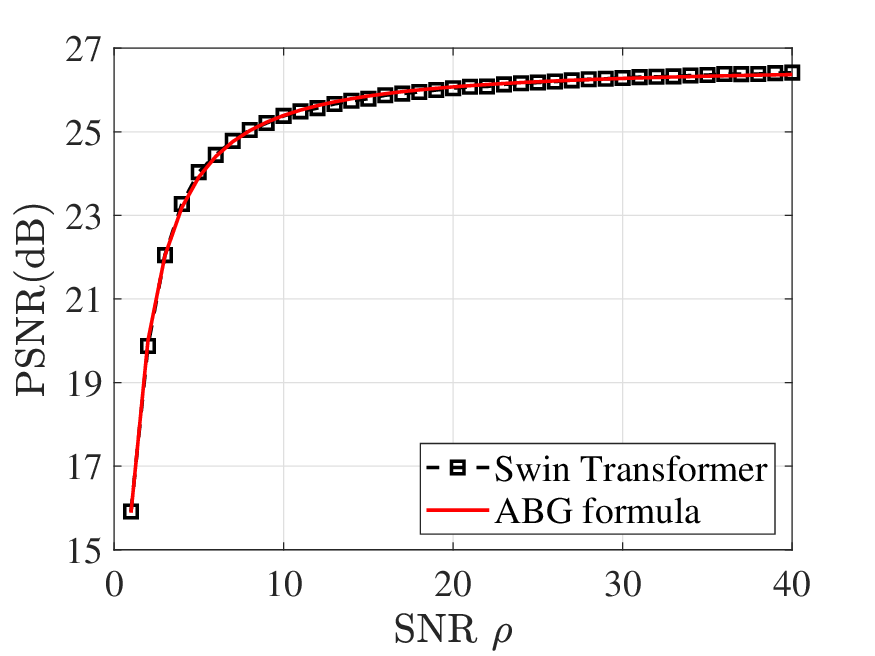}}
		\vskip-0.1cm
		\centerline{(d)}
	\end{minipage}
	\captionsetup{justification=justified}
	\caption{{PSNR performance and the corresponding ABG formulas versus SNR $\rho$ in CIFAR-10 dataset under Rayleigh channels: (a) CNN based semantic communication systems; (b) SCUNet based semantic communication systems; (c) Vision Transformer based semantic communication systems; (d) Swin Transformer based semantic communication systems.}}
	\label{ABG_PSNR_4000bit_3}
\end{figure}

\begin{figure}[!t]
	\begin{minipage}{0.25\textwidth}
		\centering
		\centerline{\includegraphics[width=\textwidth]{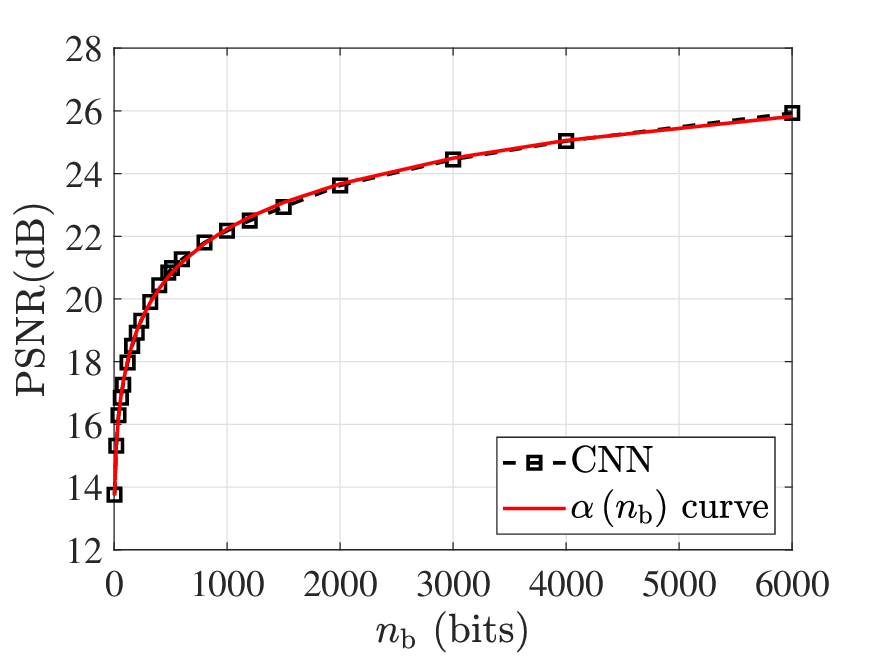}}
		\vskip - 0.1cm
		\centerline{(a)}
	\end{minipage}%
	\begin{minipage}{0.25\textwidth}
		\centering
		\centerline{\includegraphics[width=\textwidth]{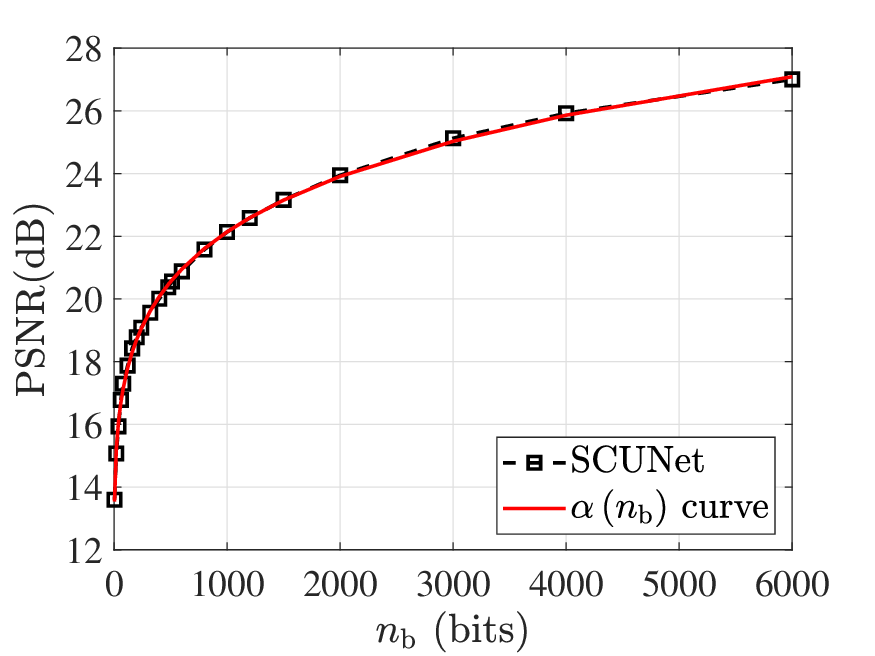}}
		\vskip - 0.1cm
		\centerline{(b)}
	\end{minipage}
	\begin{minipage}{0.25\textwidth}
		\centering
		\centerline{\includegraphics[width=\textwidth]{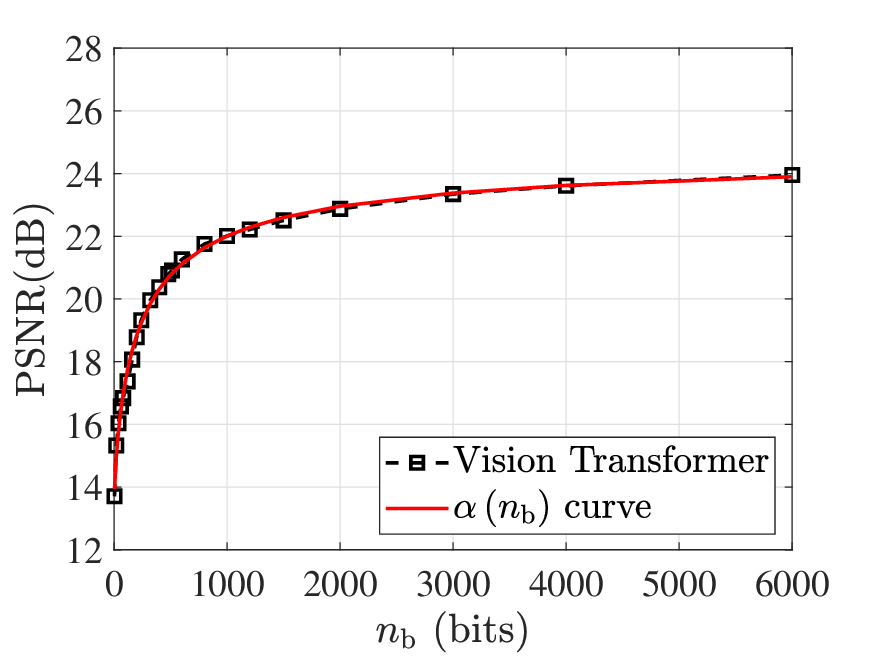}}
		\vskip - 0.1cm
		\centerline{(c)}
	\end{minipage}%
	\begin{minipage}{0.25\textwidth}
		\centering
		\centerline{\includegraphics[width=\textwidth]{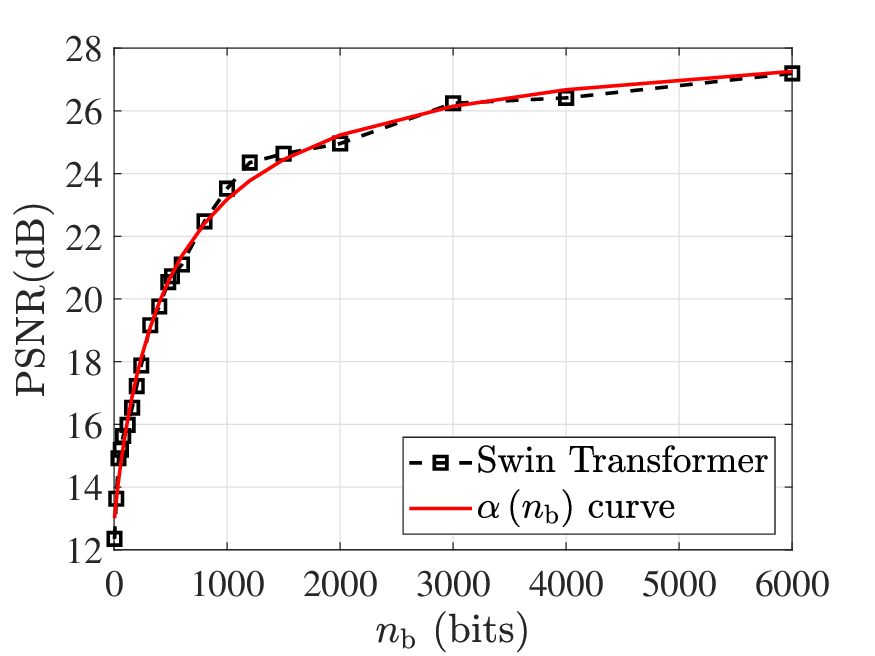}}
		\vskip - 0.1cm
		\centerline{(d)}
	\end{minipage}
	\captionsetup{justification=justified}
	\caption{{PSNR performance of semantic communication systems versus the number of quantized bits ${n_{\mathrm{b}}}$ in CIFAR-10 dataset under Rayleigh channels: (a) CNN based semantic communication systems; (b) SCUNet based semantic communication systems; (c) Vision Transformer based semantic communication systems; (d) Swin Transformer based semantic communication systems.}}
	\label{ABG_PSNR_bit_3}
\end{figure}


\begin{figure}
	\begin{minipage}{0.25\textwidth}
			\centering
			\centerline{\includegraphics[width=\textwidth]{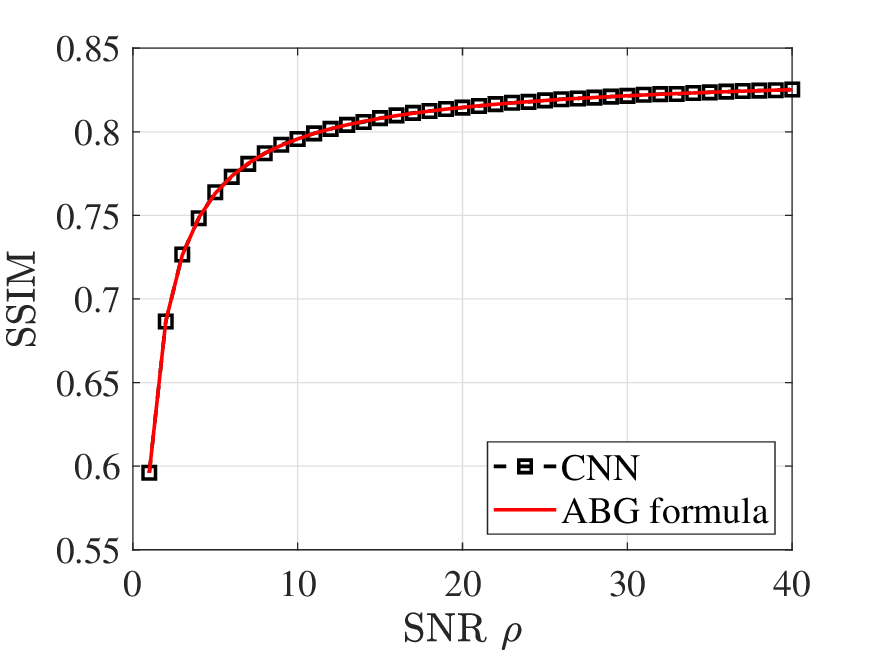}}
			\vskip - 0.1cm
			\centerline{(a)}
		\end{minipage}%
	\begin{minipage}{0.25\textwidth}
			\centering
			\centerline{\includegraphics[width=\textwidth]{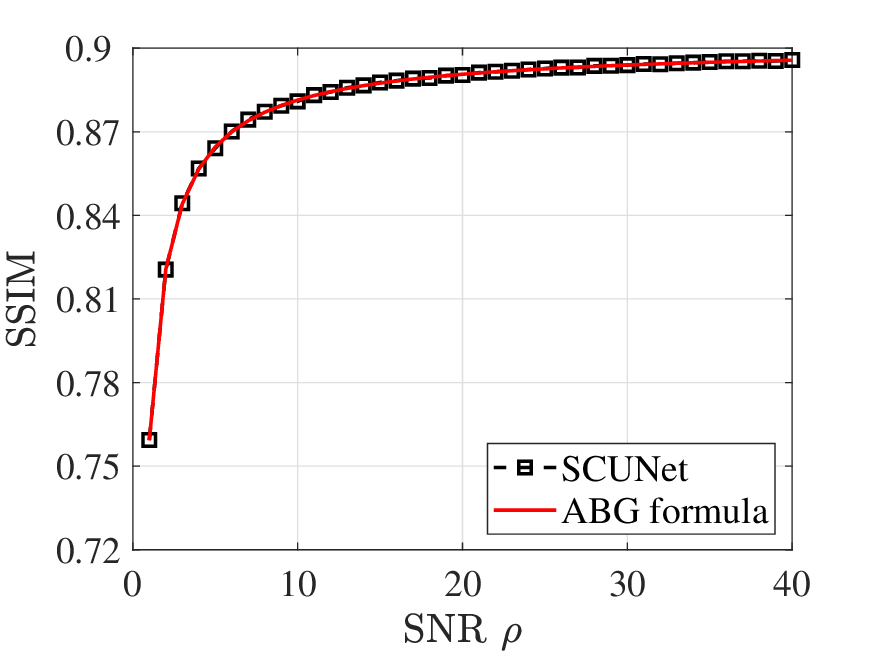}}
			\vskip - 0.1cm
			\centerline{(b)}
		\end{minipage}
	\begin{minipage}{0.25\textwidth}
			\centering
			\centerline{\includegraphics[width=\textwidth]{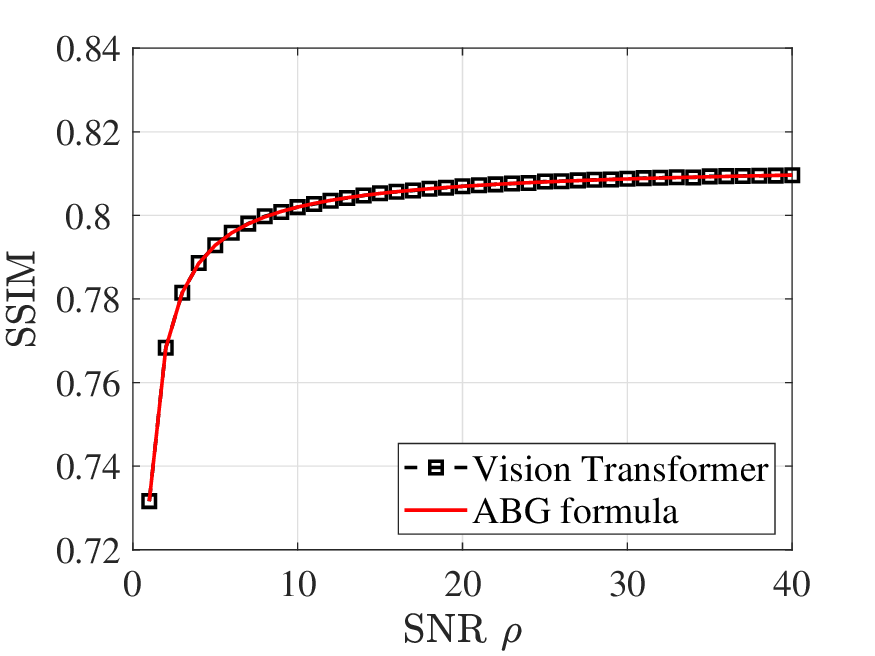}}
			\vskip - 0.1cm
			\centerline{(c)}
		\end{minipage}%
	\begin{minipage}{0.25\textwidth}
			\centering
			\centerline{\includegraphics[width=\textwidth]{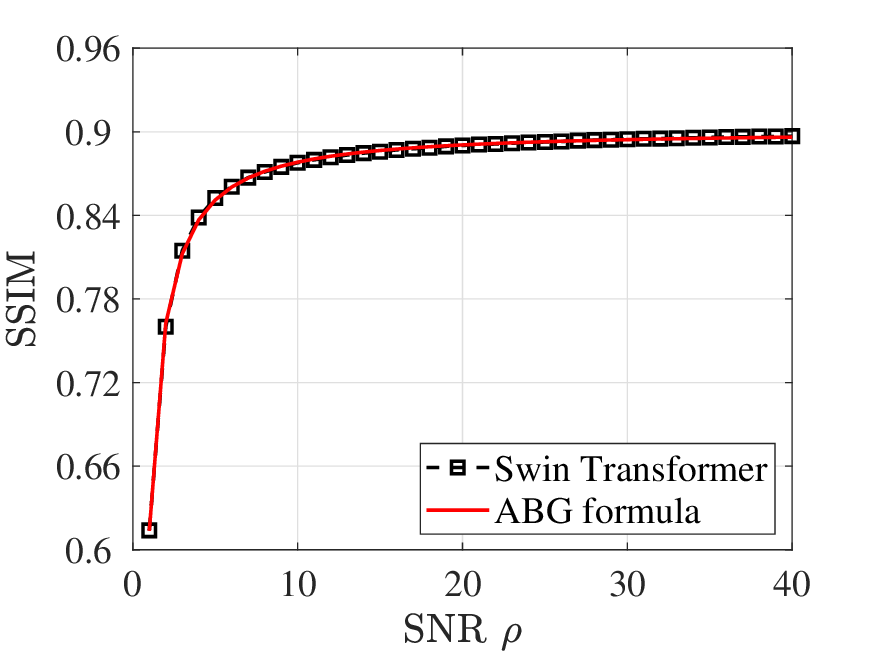}}
			\vskip - 0.1cm
			\centerline{(d)}
		\end{minipage}
	\captionsetup{justification=justified}
	\caption{{SSIM performance and the corresponding ABG formulas versus SNR $\rho$ in CIFAR-10 dataset  under Rayleigh channels: (a) CNN based semantic communication systems; (b) SCUNet based semantic communication systems; (c) Vision Transformer based semantic communication systems; (d) Swin Transformer based semantic communication systems.}}
	\label{ABG_SSIM}
\end{figure}

\begin{figure}
	\begin{minipage}{0.25\textwidth}
		\centering
		\centerline{\includegraphics[width=\textwidth]{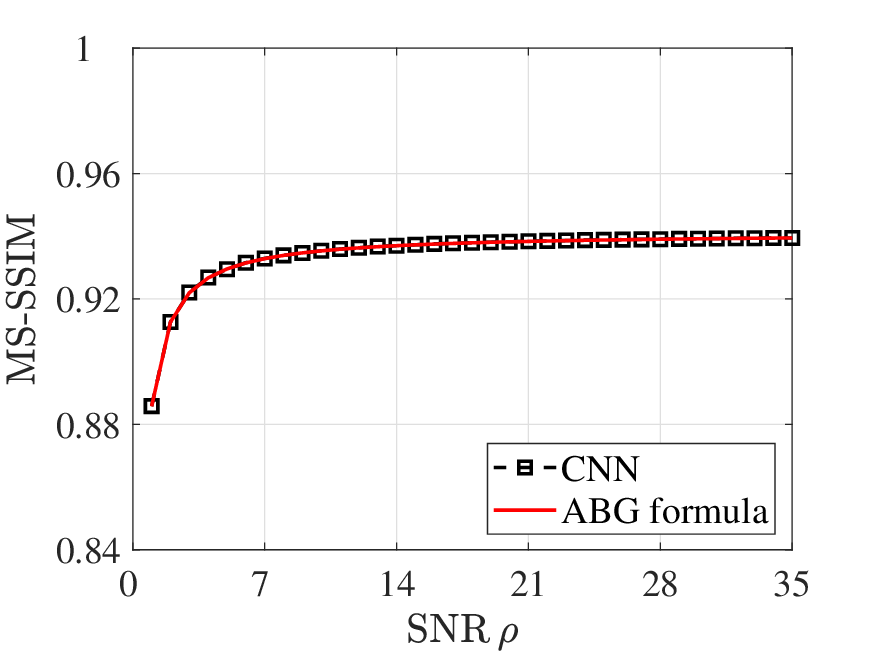}}
		\vskip - 0.1cm
		\centerline{(a)}
	\end{minipage}%
	\begin{minipage}{0.25\textwidth}
		\centering
		\centerline{\includegraphics[width=\textwidth]{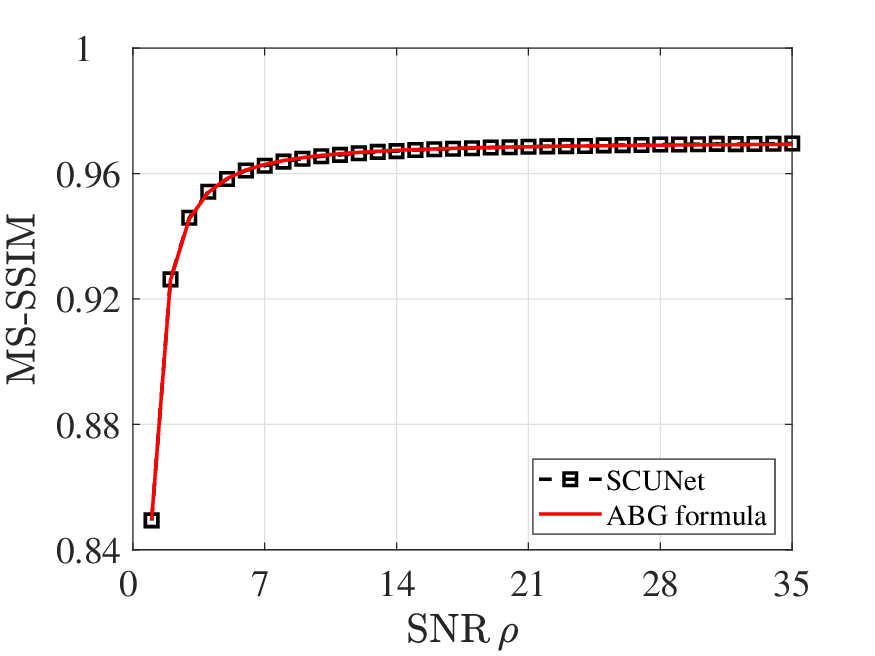}}
		\vskip - 0.1cm
		\centerline{(b)}
	\end{minipage}
	\begin{minipage}{0.25\textwidth}
		\centering
		\centerline{\includegraphics[width=\textwidth]{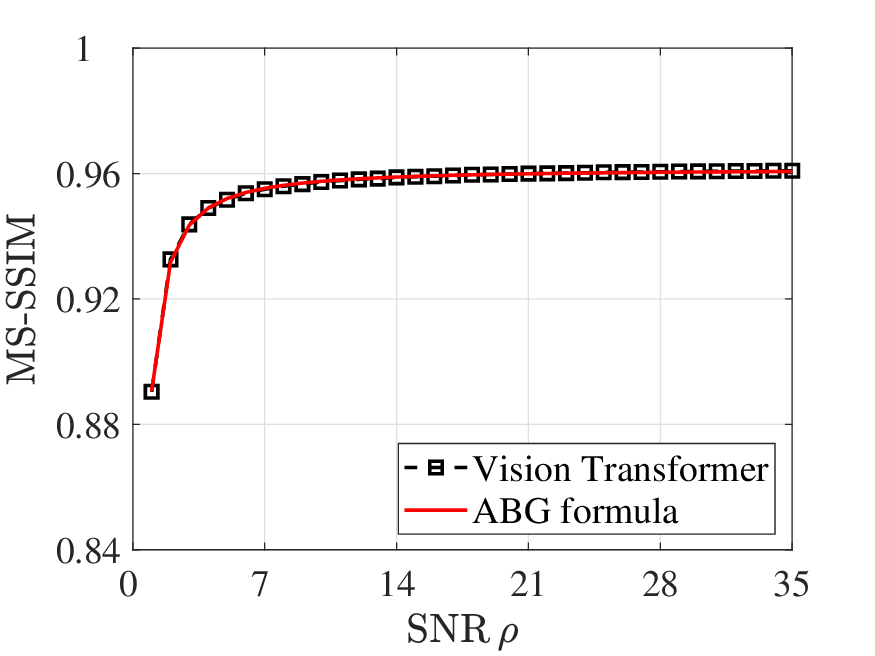}}
		\vskip - 0.1cm
		\centerline{(c)}
	\end{minipage}%
	\begin{minipage}{0.25\textwidth}
		\centering
		\centerline{\includegraphics[width=\textwidth]{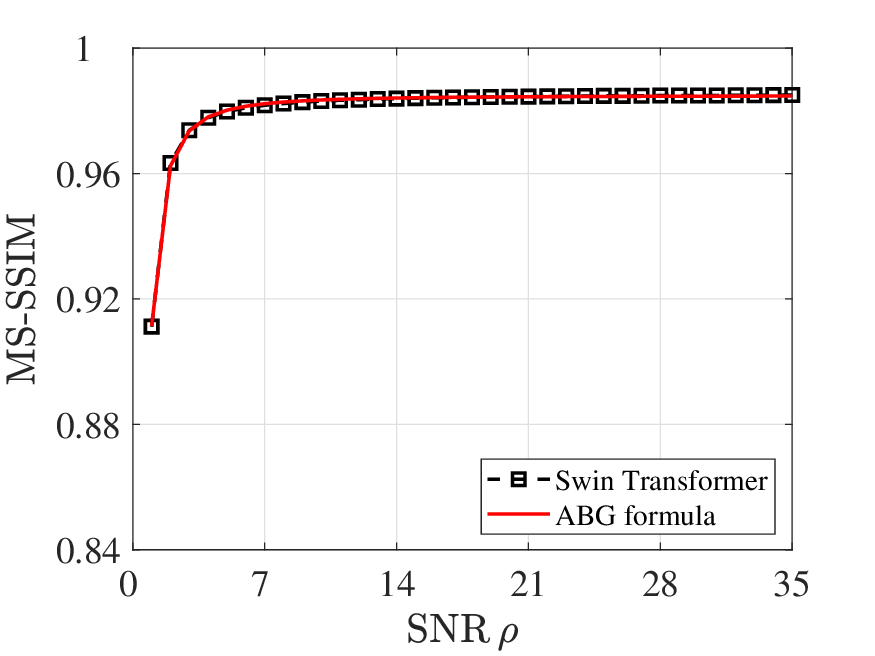}}
		\vskip - 0.1cm
		\centerline{(d)}
	\end{minipage}
	\captionsetup{justification=justified}
	\caption{{MS-SSIM performance and the corresponding ABG formulas versus SNR $\rho$ in CIFAR-10 dataset under AWGN channels: (a) CNN based semantic communication systems; (b) SCUNet based semantic communication systems; (c) Vision Transformer based semantic communication systems; (d) Swin Transformer based semantic communication systems.}}
	\label{ABG_MSSSIM_AWGN}
\end{figure}

Fig. \ref{ABG_PSNR_4000bit_3}(a), (b), (c) and (d) illustrate the PSNR performance of CNN, SCUNet, Vision Transformer and Swin Transformer based semantic communication system versus SNR under Rayleigh channels, where the black dashed lines with square markers represent the performance curves of the corresponding systems and the red solid lines denote the ABG formula fitting curves.
Meanwhile, we evaluate the accuracy of the PSNR upper bound fitted by the parameter $\alpha(n_{\mathrm{b}})$ in \eqref{upper_bound_MSSSIM}. 
Fig. \ref{ABG_PSNR_bit_3}(a), (b), (c) and (d) demonstrate that $\alpha(n_{\mathrm{b}})$ can well fit the trend of the PSNR performance upper bound as the number of quantization bits $n_{\mathrm{b}}$ changes for different semantic communication systems under Rayleigh channels.

Moreover, the ABG formula can also empirically model the relationship between end-to-end SSIM performance and SNR in semantic communication systems. As illustrated in Fig. \ref{ABG_SSIM}, the ABG formula achieves a good fitting performance  for the SSIM  curves of semantic communication models based on CNN, SCUNet, Swin Transformer, and Vision Transformer as SNR varies.

In addition to modeling the relationship between the  semantic communication performance and SNR in Rayleigh fading channels, the ABG formula can also be applied in AWGN channels.
As shown in Fig. \ref{ABG_MSSSIM_AWGN}, the ABG formula provides an accurate fit to the relationship between MS-SSIM performance and SNR for CNN, SCUNet, Swin Transformer, and Vision Transformer based semantic communication systems under AWGN channels, demonstrating its applicability to different channels.

In addition, the proposed ABG formula can also be applied to semantic communication systems with inference tasks. 
Let $\Phi \left( \cdot \right)$ denote the ABG fitting formula for the inference accuracy of semantic communication systems.
The  relationship between   inference accuracy and SNR  can be  fitted by the ABG  function as follows:
\begin{align}
   	\Phi \left( {\rho ,{n_{\mathrm{b}}}} \right) = {\alpha_2 \left( {{n_{\mathrm{b}}}} \right)} - \frac{{\gamma_2} }{{1 + {{\left( {{\beta_2} \rho } \right)}^{{\tau_2} }}}}, \label{ABG}
\end{align}
where  $ {\alpha_2 \left( {{n_{\mathrm{b}}}} \right)}> 0$ denotes the  upper bound  of the  inference accuracy, the  parameters  ${\beta_2} > 0$, ${\gamma_2} > 0$ and ${\tau_2}$ depend on the semantic encoder and decoder networks,  $\rho$ denotes the transmission SNR.

Moreover,  the   upper bound of inference accuracy ${\alpha_2 \left( {{n_{\mathrm{b}}}} \right)} $ depends on the number of  quantized output bits of     the semantic encoder $n_{\mathrm{b}}$, which is given as
\begin{align}
{ \alpha_2 \left( {{n_{\mathrm{b}}}} \right)} = {c_5} - \frac{{{c_7}}}{{1 + {{\left( {{c_6}{n_{\rm{b}}}} \right)}^{{c_8}}}}} \label{upper_bound},
\end{align}
and the parameters  ${c_5} > 0$, ${c_6} > 0$, ${c_7} > 0$,  and ${c_8}> 0$ depend on the   semantic encoder and decoder networks.
The numerical simulation of the ABG formula for inference tasks are omitted since the   performance is similar to that for the image reconstruction task as in   Section III. B.

\begin{figure}[!t]
	\centering
	\begin{minipage}[t]{0.5\textwidth}
		\centering
		\includegraphics[width=0.7\textwidth]{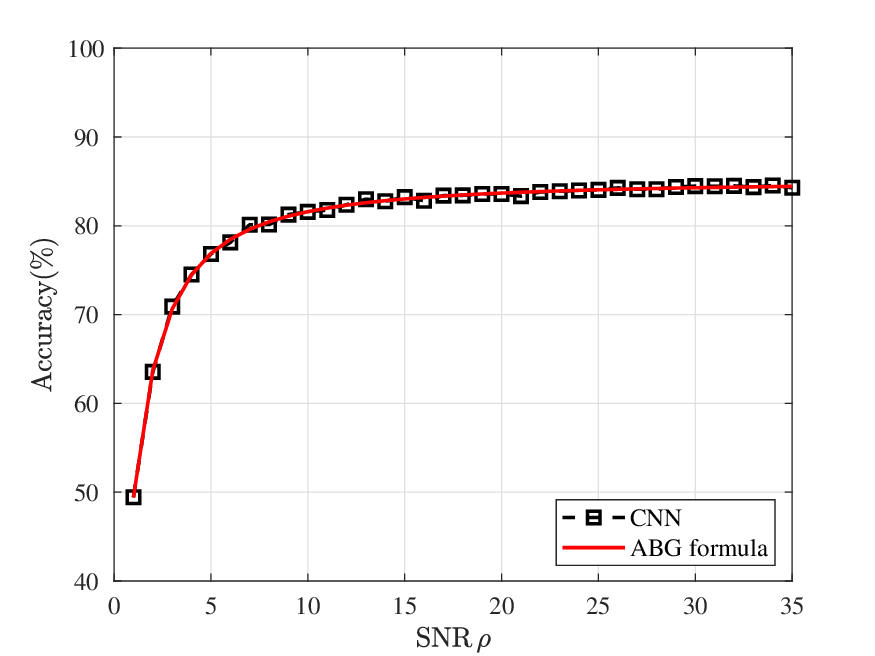} %
		\vskip-0.1cm
		\centering {(a)}
	\end{minipage}
	
	\begin{minipage}[t]{0.5\textwidth}
		\centering
		\includegraphics[width=0.7\textwidth]{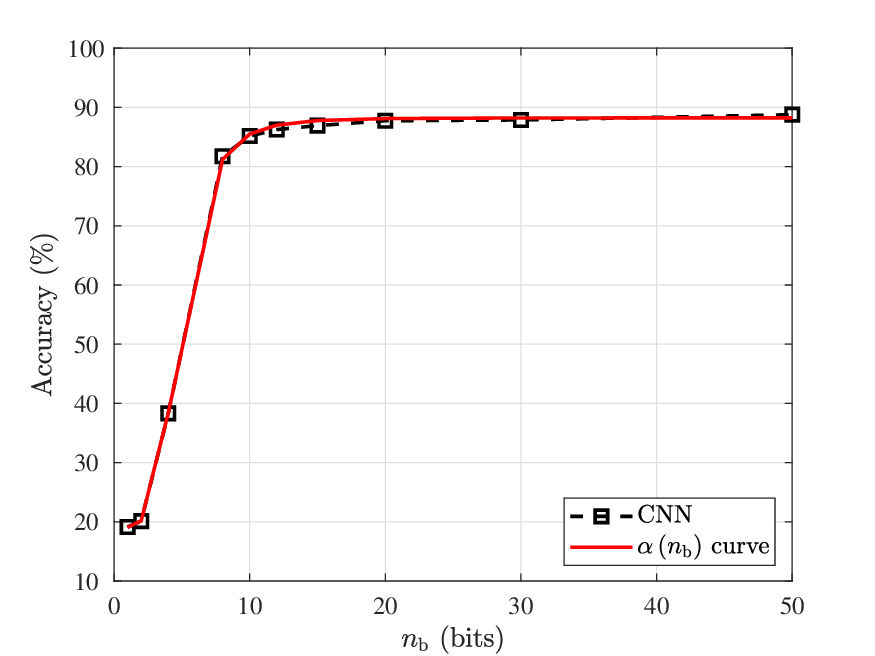} 
		\vskip-0.1cm
		\centering {(b)}
	\end{minipage}
	\captionsetup{justification=justified}
	\caption{The ABG formula is verified in the inference tasks by CNN model in CIFAR-10 dataset under Rayleigh channels: (a) Fitting inference accuracy curves with SNR $\rho$ varies; (b) Fitting inference accuracy curves with with Bits $n_{\mathrm{b}}$ varies. }
	\label{ACC_ABG1}
\end{figure}

\begin{table}[!t]
	\captionsetup{justification=justified}
	\caption{{ The parameters of  ABG formula and $\alpha_2 \left( {{n_{\mathrm{b}}}} \right)$ for fitting the CNN based  semantic communication systems for inference tasks in CIFAR-10 dataset under Rayleigh channels, and the corresponding SSE fitting errors.}}
	\centering
	\resizebox{0.9\linewidth}{!}{%
		\begin{tabular}{|c|c|c|c|c|c|}
			\hline
			\rule{0pt}{8pt}Model &  $ {\alpha_2 \left( {{n_{\mathrm{b}}}} \right) }$ & ${\gamma_2}$  &  $\beta_2$ & $\tau_2$ & Fitting errors $\varsigma_3$  \\ \hline
			\rule{0pt}{7.5pt} CNN &  { 0.85} & { 0.63} & {0.83} & {1.33}& {$1.36 \times 10^{\textsuperscript{-4}}$} \\ \hline
			\rule{0pt}{8pt}Model &  $ c_5 $ &  $c_6$ &  $c_7$ & $c_8$ & fitting errors $\varsigma_2$  \\ \hline			
			\rule{0pt}{7.5pt} CNN & { 0.88} & { 0.69}  & {0.20} & {4.50} &  {$1.61 \times 10^{\textsuperscript{-4}}$} \\ \hline	
	\end{tabular}}
	\label{fitting_error}
\end{table}


\begin{figure}[!t]
	\begin{minipage}{0.5\textwidth}
		\centerline{\includegraphics[width=0.7\textwidth]{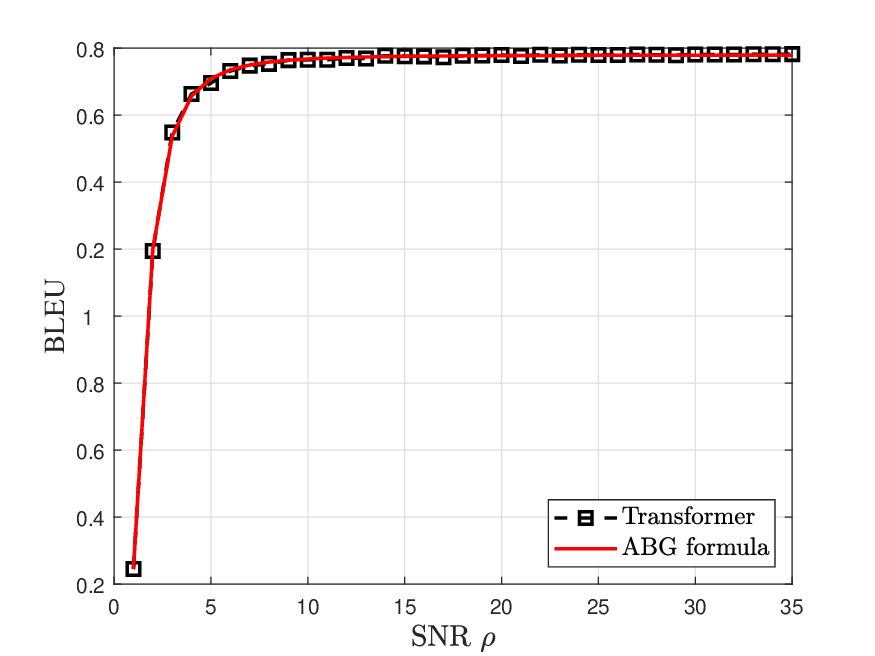}}
		\vskip-0.1cm
		\centerline{(a) }
	\end{minipage}
	
	\begin{minipage}{0.5\textwidth}
		\centerline{\includegraphics[width=0.7\textwidth]{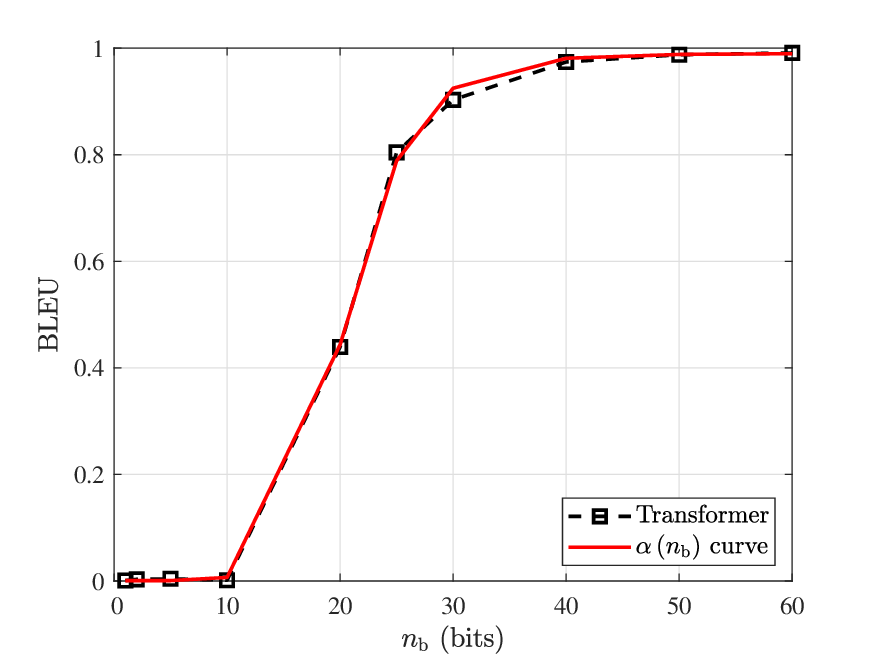}}
		\vskip-0.1cm
		\centerline{(b)}
	\end{minipage}
	\captionsetup{justification=justified}
	\caption{{	The ABG formula is verified in the text reconstruction tasks by Transformer model  under AWGN channels: (a) BLEU performance and the corresponding ABG formulas versus SNR $\rho$; (b) BLEU performance and the corresponding ABG formulas versus number of quantization bits ${n_{\mathrm{b}}}$.}}
	\label{ABG_txt_3}
\end{figure}


Fig. \ref{ACC_ABG1} (a) and (b) demonstrate the accuracy performance of CNN based semantic communication system for inference tasks versus SNR under AWGN channels, respectively, where the dashed lines with squares are the  performance curves of the corresponding semantic encoding   networks, and the solid lines are the   curves  of the  ABG formula. 
As shown in Fig. \ref{ACC_ABG1} (a) and (b) the ABG formula   can well fit the  performance curves of CNN based  semantic communication systems for inference tasks. Moreover,  with increasing SNR, the accuracy performance first rises rapidly, and then approaches the performance upper bound $ \alpha \left( {{n_{\mathrm{b}}}} \right)$.
Table \ref{fitting_error} shows the fitting parameters of the ABG formula and the performance upper bound $\alpha(n_{\mathrm{b}})$ under Rayleigh channels, with the goodness of fit evaluated using the SSE.

Specifically, for  text  reconstruction tasks, we adopt BLEU to measure the text transmission quality of the   semantic communication system.
With the  European Parliament Proceedings Parallel Corpus 1996-2011 dataset, we design semantic encoder based on  Transformer model.
Fig.\ref{ABG_txt_3} (a) and (b) show the BLEU performance versus SNR $\rho$ along with the corresponding ABG formula fitting curve, and the upper bound of BLEU performance versus the number of quantization bits $n_{b}$, respectively, along with the ABG formula fitting curve.
As shown in Fig.\ref{ABG_txt_3}, the ABG formula can well fit the BLEU performance curves of the Transformer based semantic communication system in text reconstruction tasks.

\section{Optimal Power Allocation for Semantic Communications}

Without a tractable system performance model, it is challenging for traditional   semantic communication systems to implement real-time  adaptive power control. Thus, it will result in the inefficient power allocation to guarantee the quality of semantic transmission, or allocating too much power, that leads to power waste and interference to other users.
To address this challenge, we investigate optimal power allocation schemes for  semantic communications systems with  single users and multi-users, respectively.

 \subsection{Adaptive Power Control for Single User Case}
Consider a single  user semantic communication system for image reconstruction tasks with time-varying fading channel ${h\left( t \right)}$, as shown in Section II. Based on the established the ABG formula, the relationship between power control and performance can be established. According to \eqref{ABG},   the  relationship between  MS-SSIM and transmitted power  $p\left( t \right)$ is  as
 \begin{align}
 	{\rm{\varphi }}\left( {p\left( t \right),{n_{\mathrm{b}}}} \right) = \alpha  - \frac{\gamma }{{1 + {{\left( {\beta \frac{{p\left( t \right){{\left| {h\left( t \right)} \right|}^2}}}{{{\sigma ^2}}}} \right)}^\tau }}}.\label{ABG_power}
\end{align}

Given a   image reconstruction  quality threshold $\eta $, i.e.,
\begin{align}
   	\alpha - \frac{\gamma }{{1 + {{\left( {\beta \frac{{p\left( t \right){{\left| {h\left( t \right)} \right|}^2}}}{{{\sigma ^2}}}} \right)}^\tau }}} \ge \eta,
\end{align}
the optimal adaptive power control  ${p^*}\left( t \right)$  is given as
 \begin{align}
 	{p^*}\left( t \right) = \frac{{{\sigma ^2}}}{{\beta {{\left| {h\left( t \right)} \right|}^2}}}{\left( {\frac{\gamma }{{\alpha \left( {{n_{\mathrm{b}}}} \right) - \eta }} - 1} \right)^{\frac{1}{\tau }}}.
\end{align}


\subsection{Energy Efficiency}
Furthermore, we investigate the energy efficiency of the    semantic communication system.
Let  $p$ and ${p_{\rm{cir}}}$ denote the transmitted power and the circuit power consumption  of the semantic communication system, respectively.
Let ${\psi _{{\rm{EE}}}}$ denote the  energy efficiency of the    semantic communication system, which is  defined as the ratio  between image reconstruction  quality and the   transmitted power, i.e.,
 \begin{align}
{\psi _{{\rm{EE}}}}\left( p \right) &\buildrel \Delta \over = \frac{{{\rm{\varphi  }}\left( p \right)}}{{p + {p_{{\rm{cir}}}}}} \nonumber\\
&= \left( {\alpha  - \frac{\gamma }{{1 + {{\left( {\beta \frac{{p{{\left| h \right|}^2}}}{{{\sigma ^2}}}} \right)}^\tau }}}} \right)\frac{{\rm{1}}}{{p + {p_{{\rm{cir}}}}}}.
\end{align}

Since  ${{\rm{\varphi }}\left( p \right)}$ is a differentiable concave function of transmitted power $p$, and the denominator is an affine function of $p$, the energy efficiency of the semantic communication system
${\psi _{{\rm{EE}}}}$ is a quasi-concave function of  $p$.

\begin{figure}
	\centering
	\includegraphics[width=0.44\textwidth]{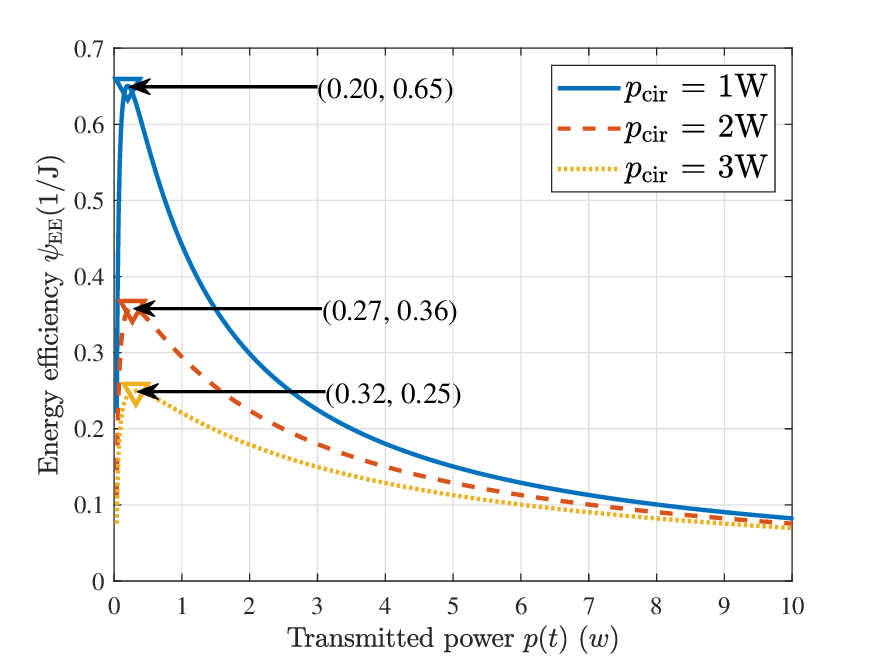}
	\captionsetup{justification=justified}
	\caption{{Energy efficiency  ${\psi _{{\rm{EE}}}}$ versus transmitted power  $p$ and the circuit power consumption ${p_{\rm{cir}}}$.}}
	\label{ee}
\end{figure}

Next, we check the energy efficiency ${\psi _{{\rm{EE}}}}$ versus transmitted power  $p$ and the circuit power consumption ${p_{\rm{cir}}}$ in Fig. \ref{ee}. It can be found that the energy efficiency ${\psi _{{\rm{EE}}}}$ first monotonically increases and then decreases with respect to $p$, which verifies that ${\psi _{{\rm{EE}}}}$ is a quasi-concave function of $p$.
Moreover, as shown in Fig. \ref{ee}, we use the triangle symbol to denote the optimal point of each circuit power consumption scenario, where the corresponding optimal power consumption and the optimal energy efficiency are also marked.
The circuit power consumption, $ P_{\text{cir}} $, is a critical energy efficiency metric in communication systems, influenced by various factors such as hardware design, component efficiency, operating frequency, and usage scenarios. 
For instance, the efficiency of key modules like power amplifiers and analog-to-digital converters significantly impacts the value of $ P_{\text{cir}} $. 
Moreover, higher operating frequencies can lead to increased dynamic power consumption, while low-power designs or optimized hardware architectures can help reduce $ P_{\text{cir}}$. In Fig. 8, different values of $ P_{\text{cir}} $ correspond to varying operational conditions, and it can be observed that as $P_{\text{cir}} $ decreases, the optimal energy efficiency gradually diminishes.

In the following, we investigate the energy efficiency maximization problem of  the semantic communication system subject to both the minimum  transmission rate requirement\footnote{For transmission delay consideration,  there is a   transmission rate constraint.} and the total power threshold constraint, which can be mathematically formulated as
\begin{subequations}\label{EE_opt}
\begin{align}
\mathop {\max }\limits_p &~{\psi _{{\rm{EE}}}}\left( p \right)\label{EE_opta}\\
{\rm{s}}{\rm{.t.}}&~{B}{\rm{lo}}{{\rm{g}}_2}\left( {1 + \frac{{p{{\left| h \right|}^2}}}{{{\sigma ^2}}}} \right) \ge \overline r, \label{EE_optb}\\
&~0 \le p \le {{P_{\rm{U}}}},\label{EE_optc}
\end{align}
\end{subequations}
where $B$ represents the bandwidth, $\overline r $ denotes the  minimum  transmission rate requirement, and ${{P_{\rm{U}}}}$  denotes the total power threshold.

Although problem \eqref{EE_opt} is non-convex and intractable, it is a concave-linear fractional problem.
To handle  problem \eqref{EE_opt}, we design a parametric method based on Dinkelbach-type algorithm \cite{Zappone_2015}. 
Specifically, problem  \eqref{EE_opt} is first reformulated into a series of auxiliary convex   sub-problems, and then by maximizing the  sub-problems in each iteration, the optimal solutions of   sub-problems    converge to the global optimal solution of problem \eqref{EE_opt}.

More specifically,  under constraints \eqref{EE_optb}  and \eqref{EE_optc}, the   feasible region of problem \eqref{EE_opt} is given as
 \begin{align}
 	{P_{\rm{L}}} \le p \le {P_{\rm{U}}},
  \end{align}
 where ${P_{\rm{L}}} \buildrel \Delta \over = \frac{{\left( {{2^{ { \frac{{\bar r}}{B} }}} - 1} \right){\sigma ^2}}}{{{{\left| h \right|}^2}}}$. Furthermore,  we introduce a new parametric function  $F\left( {p,v} \right)$ defined as
 \begin{align}
 	F\left( {p,v} \right) \buildrel \Delta \over = \alpha  - \frac{\gamma }{{1 + {{\left( {\beta \frac{{p{{\left| h \right|}^2}}}{{{\sigma ^2}}}} \right)}^\tau }}} - \upsilon\left( {p + {p_{{\rm{cir}}}}} \right),
 \end{align}
where $\upsilon \ge 0$ is an auxiliary variable.

 Given
some $v$, we obtain the     sub-problem    as follows:
  \begin{subequations}\label{sub_c}
  \begin{align}
\mathop {\max }\limits_p &~F\left( {p,\upsilon} \right)\\
{\rm{s}}{\rm{.t}}{\rm{.}}&~{P_{\rm{L}}} \le p \le {P_{\rm{U}}}.\label{sub_c1}
 \end{align}
   \end{subequations}

Then, the optimal   solutions of problem \eqref{EE_opt} can be obtained
by solving \eqref{sub_c} with the largest feasible value of $v$ achievable.

Note that, problem \eqref{sub_c} is convex with respect to $p$.
Thus, by    taking the partial derivative of the objective function  $F\left( {p,\upsilon} \right)$  to be zero, i.e.,
  $\frac{{\partial F\left( {p,\upsilon} \right)}}{{\partial p}} = 0$,
 we obtain
 \begin{align}\label{equ_c}
{\left( {\beta \frac{{p{{\left| h \right|}^2}}}{{{\sigma ^2}}}} \right)^{\frac{{\tau  - 1}}{2}}} = \sqrt {\frac{{\upsilon{\sigma ^2}}}{{\gamma \tau \beta {{\left| h \right|}^2}}}} \left( {1 + {{\left( {\beta \frac{{p{{\left| h \right|}^2}}}{{{\sigma ^2}}}} \right)}^\tau }} \right).
 \end{align}

Let $\widetilde p$ denote the roots of equation \eqref{equ_c}. Since \eqref{equ_c} is a nonlinear equation, $\widetilde p$ can be efficiently solved by using off-the-shelf
nonlinear optimization solvers.

Let ${p^*}$ denote   the optimal solution of problem \eqref{sub_c}, which is given as
\begin{align}\label{opt_sub}
{p^*} = {\rm{Pro}}{{\rm{j}}_\Omega }\left( {\tilde p} \right) = \left\{ \begin{array}{*{20}{l}}
		{{P_{\rm{L}}},\;\;\tilde p \le {P_{\rm{L}}},}\\
		{\tilde p,\;\;{P_{\rm{L}}} \le \tilde p < {P_{\rm{U}}},}\\
		{{P_{\rm{U}}},\;\;\tilde p \ge {P_{\rm{U}}},}
\end{array} \right.
\end{align}
where $\Omega  \buildrel \Delta \over = [{P_{\rm{L}}},{P_{\rm{U}}}]$ and ${\rm{Pro}}{{\rm{j}}_\Omega }\left( {\tilde p} \right)$ denotes the projection of the point ${\tilde p}$ onto the space $\Omega $.

Finally, the energy efficiency maximization problem \eqref{EE_opt} can
be solved by the Dinkelbach-type algorithm, and the details
of implementation of the algorithm are given in Algorithm \ref{algorithm_DT}. 
The Dinkelbach-type algorithm is guaranteed to converge to the optimal
solution of \eqref{EE_opt} in a finite number of iterations \cite{Matlab_2012}.

\begin{algorithm}
	\caption{  Dinkelbach-Type Algorithm for problem \eqref{EE_opt} }
	\label{algorithm_DT}
	\begin{algorithmic}[1]
		\State {\bf  Initialization:} Given a termination parameter ${\xi   \ge 0 }$, set   ${n=0}$ and ${\upsilon _n} = 0$;
		\State {\bf  repeat:}
		\State \hspace*{0.2in}  ${n \leftarrow n + 1}$.
		\State \hspace*{0.2in}  Calculate  the  solution  ${p^*}$ based on \eqref{opt_sub};
		\State \hspace*{0.2in}  Update ${\upsilon _n} = { {\psi _{{\rm{EE}}}}\left( {{\rm{p}^*}} \right)}$;
 		\State {\bf  until:} $F\left( {{p^*},{\upsilon _n}} \right) \le \xi $;
		\State {\bf  Output}: The optimal solution ${{\bf{p}}^*}$, and the maximum energy efficiency {${\psi _{{\rm{EE}}}}\left( {{\rm{p}^*}} \right)$}.
	\end{algorithmic}
\end{algorithm}




\subsection{Optimal  Power Allocation for Multi-user Case  }

Consider a typical multi-user OFDMA downlink semantic communication system for image reconstruction tasks, in which the base station (BS) transmits  image to $ K$ users through $K$ orthogonal subcarriers, and each user is allocated one subcarrier. Without loss of generality, assume that the $i$th subcarrier  is allocated to user $i$, $i=1,...,K$. Let ${h_i}$ denote the channel gain of between BS   and user $i$. The MS-SSIM of the image reconstruction    of user $i$ is given as
 \begin{align}
 	{{\rm{\varphi }}_i}\left( {{p_i}} \right) = {\alpha _i} - \frac{{{\gamma _i}}}{{1 + {{\left( {{\beta _i}\frac{{{p_i}{{\left| {{h_i}} \right|}^2}}}{{\sigma _i^2}}} \right)}^{{\tau _i}}}}},    
 \end{align}
where   ${p_i}$ denote the allocated power of user $i$.
Let $\bar P$ denote the total transmit power of the semantic BS.

For the multi-user OFDMA downlink semantic communication system,  we aim to optimize the  power allocation to
  maximize the minimum (worst case)   MS-SSIM of $K$ users subject
to the total power constraint,  which can be mathematically formulated as
 \begin{subequations}\label{max_min}
	\begin{align}
		\mathop {\max }\limits_{\{ {p_i}\} _{i = 1}^K} &\mathop {\min }\limits_{i = 1,...,K}  {{{\rm{\varphi }}_i}\left( {{p_i}} \right)} \\
		{\rm{s}}{\rm{.t}} &~ {\rm{.}}\sum\limits_{i = 1}^K {{p_i}}  \le \bar P,{p_i} \ge 0,\;i = 1,...,K.
	\end{align}
\end{subequations}

%

Due to non-convexity of the objective function \eqref{max_min}, problem (18) is a difficult problem.
To address this difficulty, we introduce an auxiliary variable $\nu \in \left[ {0,1 } \right]$, and  equivalently reformulate  problem \eqref{max_min}   as
 \begin{subequations}\label{max}
	\begin{align}
	\mathop {\max } \limits_{\{ {p_i}\} _{i = 1}^K}&~\;\nu\\
	{\rm{s}}{\rm{.t}}{\rm{.}}&~{{\rm{\varphi}}_i}\left( {{p_i}} \right) \ge \nu,i = 1,...,K\\
	&~\sum\limits_{i = 1}^K {{p_i}}  \le \bar P,{p_i} \ge 0,\;i = 1,...,K.
	\end{align}
\end{subequations}

For any fixed $\nu \in \left[ {0,1 } \right]$,  problem \eqref{max_2} is
a linear programming problem. Thus, problem \eqref{max_2}  is quasi-convex, and we can find
the globally optimal solutions by a simple bisection method.
Specifically, problem  \eqref{max_2}  is transformed into a series of convex
feasibility subproblems, for a given value $\nu \in \left[ {0,1 } \right]$:
 \begin{subequations}\label{max_2}
	\begin{align}
	{\rm{find}}&~\left\{ {{p_i}} \right\}_{i = 1}^K\\
	{\rm{s}}.{\rm{t}}.&~\frac{{{p_i}{{\left| {{h_i}} \right|}^2}}}{{{\sigma _i}^2}} \ge \frac{1}{{{\beta _i}}}{\left( {\frac{{{\gamma _i}}}{{{\alpha _i} - \nu }} - 1} \right)^{\frac{1}{\tau_i }}},i = 1,...,K,\\
	&~\sum\limits_{i = 1}^K {{p_i}}  \le \bar P,{p_i} \ge 0,\;i = 1,...,K.
	\end{align}
\end{subequations}

Problem \eqref{max_2} is a linear programming problem, and can be efficiently solved by the simplex method\cite{Litchfield_JPET_1949}.

To find the optimal power allocation ${\left\{ {{p_i}} \right\}_{i = 1}^K}$ of problem \eqref{max_2}, we adopt a bisection search method, and the details are given in   Algorithm~\ref{table2}.

\begin{algorithm}[!t]
	\caption{Bisection  algorithm for problem \eqref{max}}
	\label{table2}
	\begin{algorithmic}
		\State {\bf  Initialization:} Given a termination parameter ${\delta  \ge 0 }$, initialize the proper parameters ${\nu _{\rm{L}}} = 0$ and ${\nu _{\rm{U}}} = 1$ such that ${\nu ^{{\rm{opt}}}} \in \left[ {{\nu _{\rm{L}}},{\nu _{\rm{U}}}} \right]$;
		\State {\bf  repeat:}
		\State \hspace*{0.2in} $\nu  = \frac{{{\nu _{\rm{L}}} + {\nu _{\rm{U}}}}}{2}$, and check the feasibility of the problem
\eqref{max_2};
		\State \hspace*{0.2in}  If   feasible solutions ${\left\{ {{p_i}} \right\}_{i = 1}^K}$ are found, set ${\nu _{\rm{L}}} = \nu $; otherwise, set ${\nu _{\rm{U}}} = \nu $;
 		\State {\bf  until:}  ${\nu _{\rm{U}}} - {\nu _{\rm{L}}} \le \delta$;
				\State {\bf Output}: The optimal power allocation ${\left\{ {{p_i}} \right\}_{i = 1}^K}$.
	\end{algorithmic}
\end{algorithm}

\section{Simulation Results and Discussions}
To illustrate the performance of the  ABG formula  based power allocation schemes proposed in Section IV, we present detailed numerical results for  semantic communication systems.

\begin{figure}
		\begin{minipage}[t]{0.5\textwidth}
			\centering
			\includegraphics[width=0.78\textwidth]{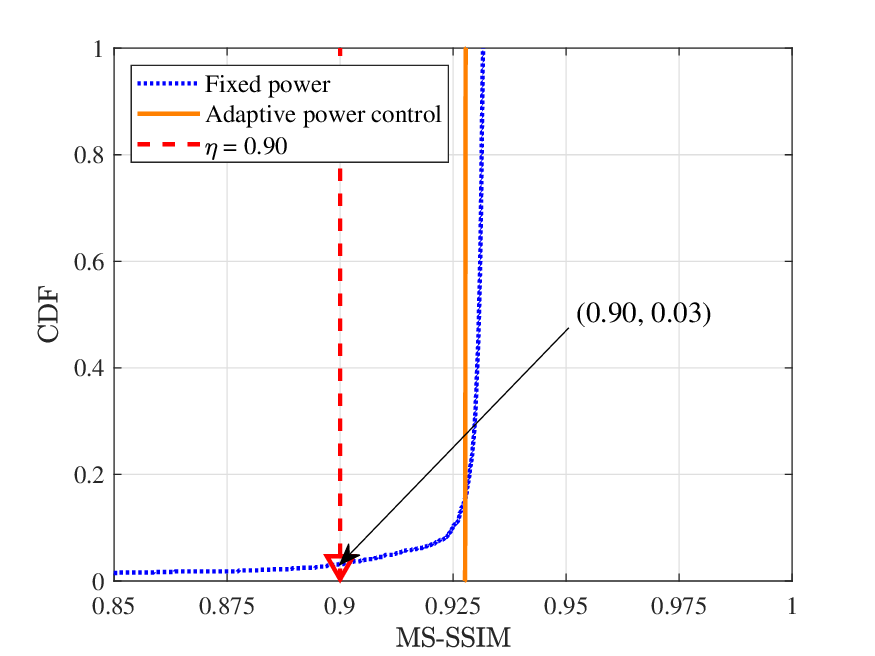}
			\vskip-0.1cm
			\centering {(a)}
		\end{minipage}
		\begin{minipage}[t]{0.5\textwidth}
			\centering
			\includegraphics[width=0.78\textwidth]{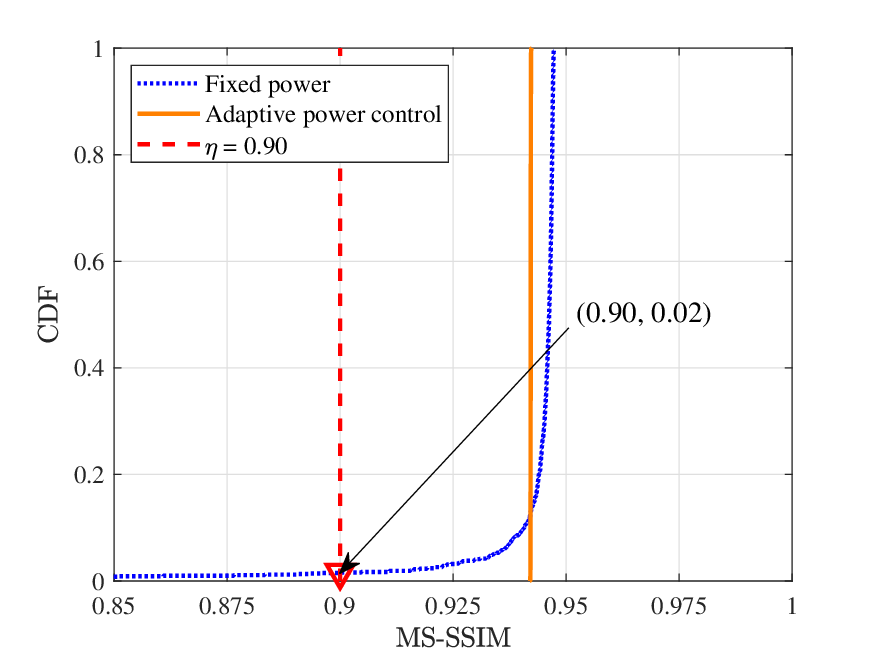}
			\vskip-0.1cm
			\centering {(b)}
		\end{minipage}
	
		\begin{minipage}[t]{0.5\textwidth}
			\centering
			\includegraphics[width=0.78\textwidth]{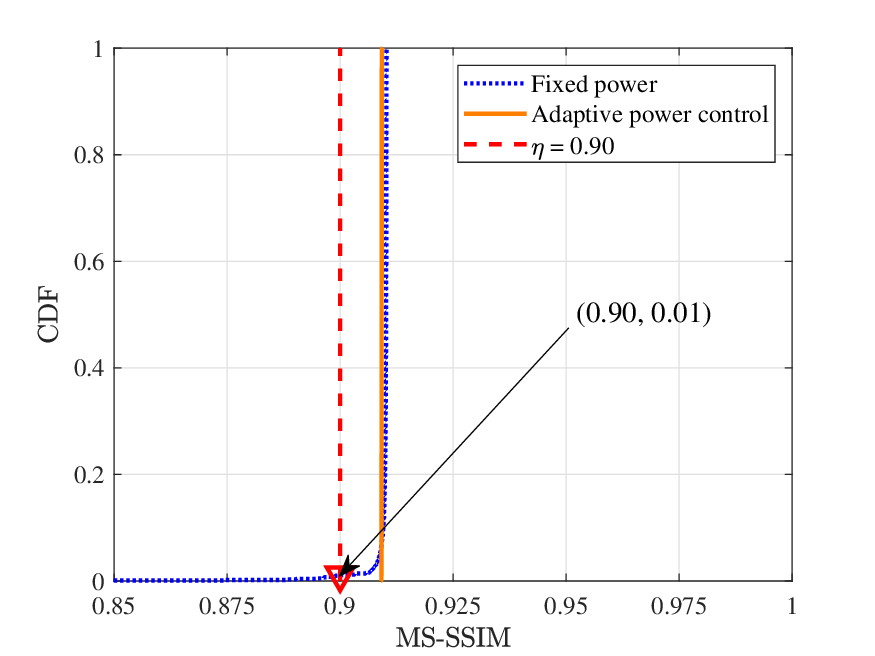}
			\vskip-0.1cm
			\centering {(c)}
		\end{minipage}
		\begin{minipage}[t]{0.5\textwidth}
			\centering
			\includegraphics[width=0.78\textwidth]{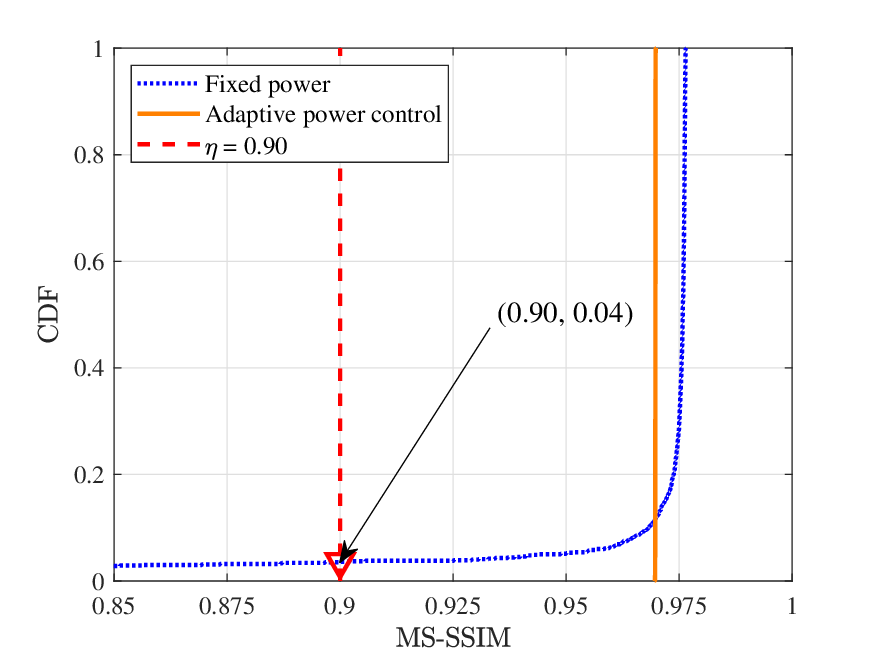}
			\vskip-0.1cm
			\centering {(d)}
		\end{minipage}
	\captionsetup{justification=justified}
	\centering
	\caption{The empirical CDF of MS-SSIM with the   given MS-SSIM threshold $\eta = 0.90$. (a) CNN  based semantic communication systems; (b)   SCUNet based semantic communication systems; (c) Vision Transformer based semantic communication systems; (d) Swin Transformer based semantic communication systems.   }
	\label{cdf_cifar10}
\end{figure}

 \subsection{Evaluation of Adaptive Power Control for Single User Case}
 In this subsection, we evaluate the proposed ABG formula  based   adaptive   power control scheme for semantic communication systems. Moreover, we present the traditional fixed  transmission power
scheme  for comparison.

%


Fig.\ref{cdf_cifar10} (a), (b), (c), (d) illustrate the   cumulative distribution functions (CDF) of MS-SSIM   of    the CNN, SCUNet, Vision Transformer, and Swin Transformer based semantic   communication systems over CIFAR-10 dataset, where the given MS-SSIM threshold  is $\eta = 0.90$.
  Each curve in the figure represents
the empirical CDF of the MS-SSIM performance obtained from 10 000
random channel realizations.
From the figures, we observe that the outage of the traditional fixed  transmission power scheme CNN, SCUNet, Vision Transformer, and Swin Transformer based semantic   communication systems are $0.03$, $0.02$, $0.01$, and $0.04$, respectively, which   violates the minimum MS-SSIM requirements.
Meanwhile, Fig.\ref{cdf_cifar10} (a), (b), (c), (d) show that the proposed adaptive power control scheme  for all four DL based  semantic   communication systems  satisfy the minimum MS-SSIM requirements.
These results show the necessity of adaptive power control, and the effectiveness of the proposed adaptive power control scheme.




\begin{figure}[!t]
	\centering
	\includegraphics[width=0.4\textwidth]{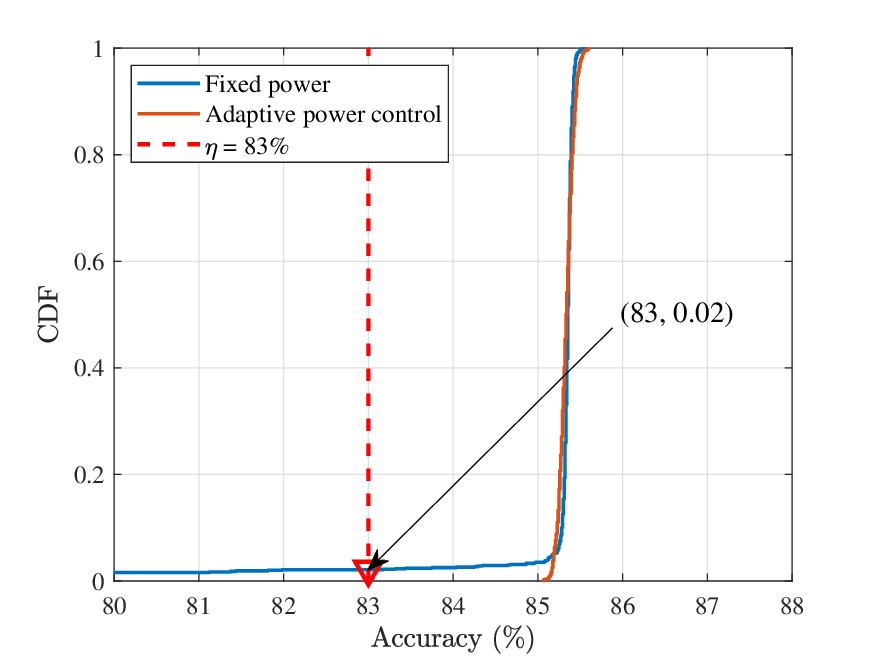}
	\captionsetup{justification=justified}
	\caption{The empirical CDF of accuracy of the fixed power scheme and adaptive power control scheme with the  inference threshold  $\eta = 83\%$.}
	\label{Inference_task}
\end{figure}

Fig. \ref{Inference_task} plots the empirical CDF of accuracy  of the fixed power scheme and the proposed adaptive power control scheme with the  inference threshold  $\eta = 83\%$.
Clearly, the inference results of  the fixed power scheme  cannot satisfy  the minimum accuracy requirements, and about $0.02$ of the accuracy are below the target threshold  $\eta = 83\%$. On the other hand, the inference results of  the   proposed adaptive power control scheme satisfy the minimum accuracy requirements.   
This result also verifies the necessity of adaptive power control, and the effectiveness of the proposed adaptive power control scheme.


\begin{figure}[!t]
	\centering
	\begin{minipage}[t]{0.5\textwidth}
		\centering
		\includegraphics[width=0.8\textwidth]{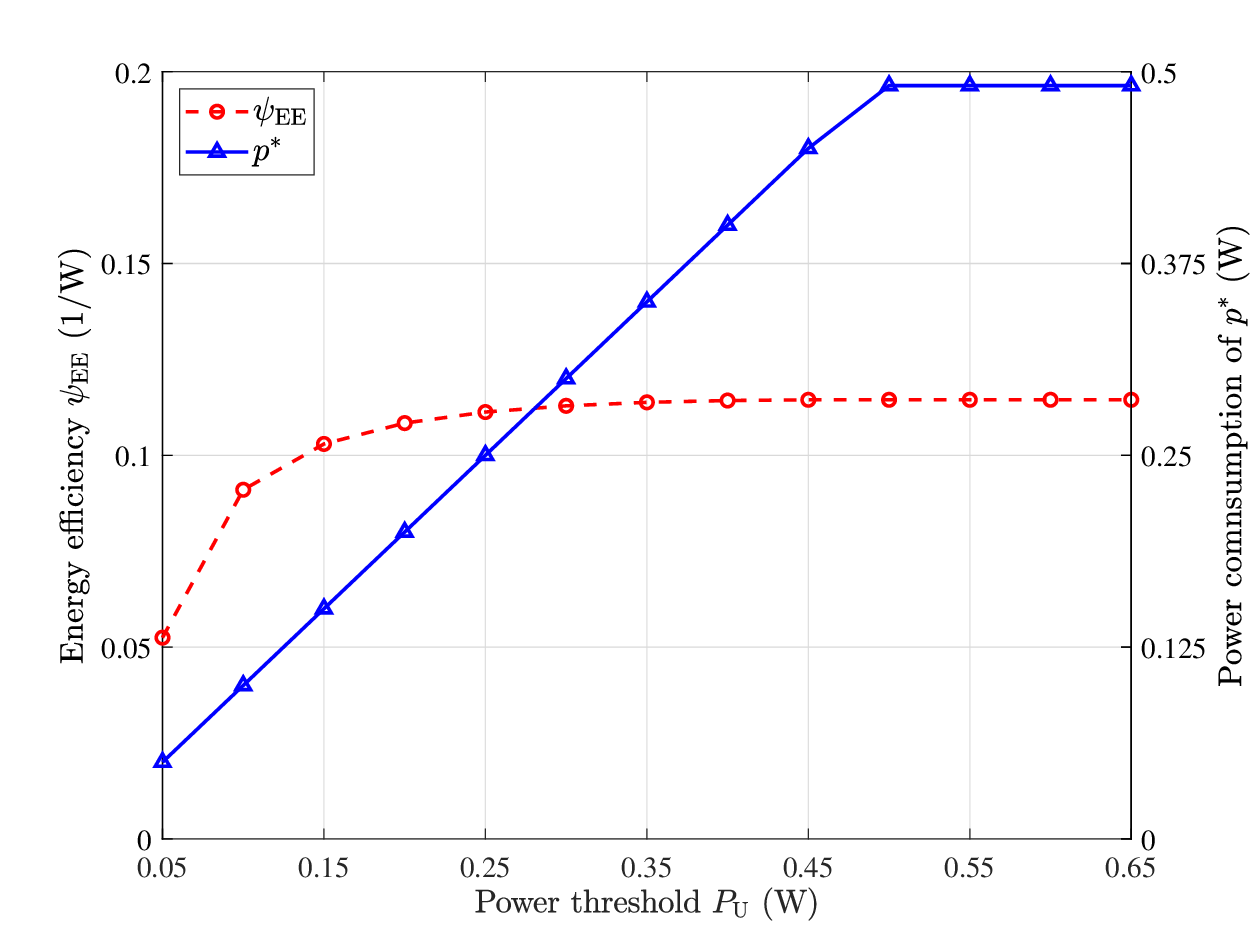} 
		\vskip-0.1cm
		\centering {(a)}
	\end{minipage}
	
	\begin{minipage}[t]{0.5\textwidth}
		\centering
		\includegraphics[width=0.8\textwidth]{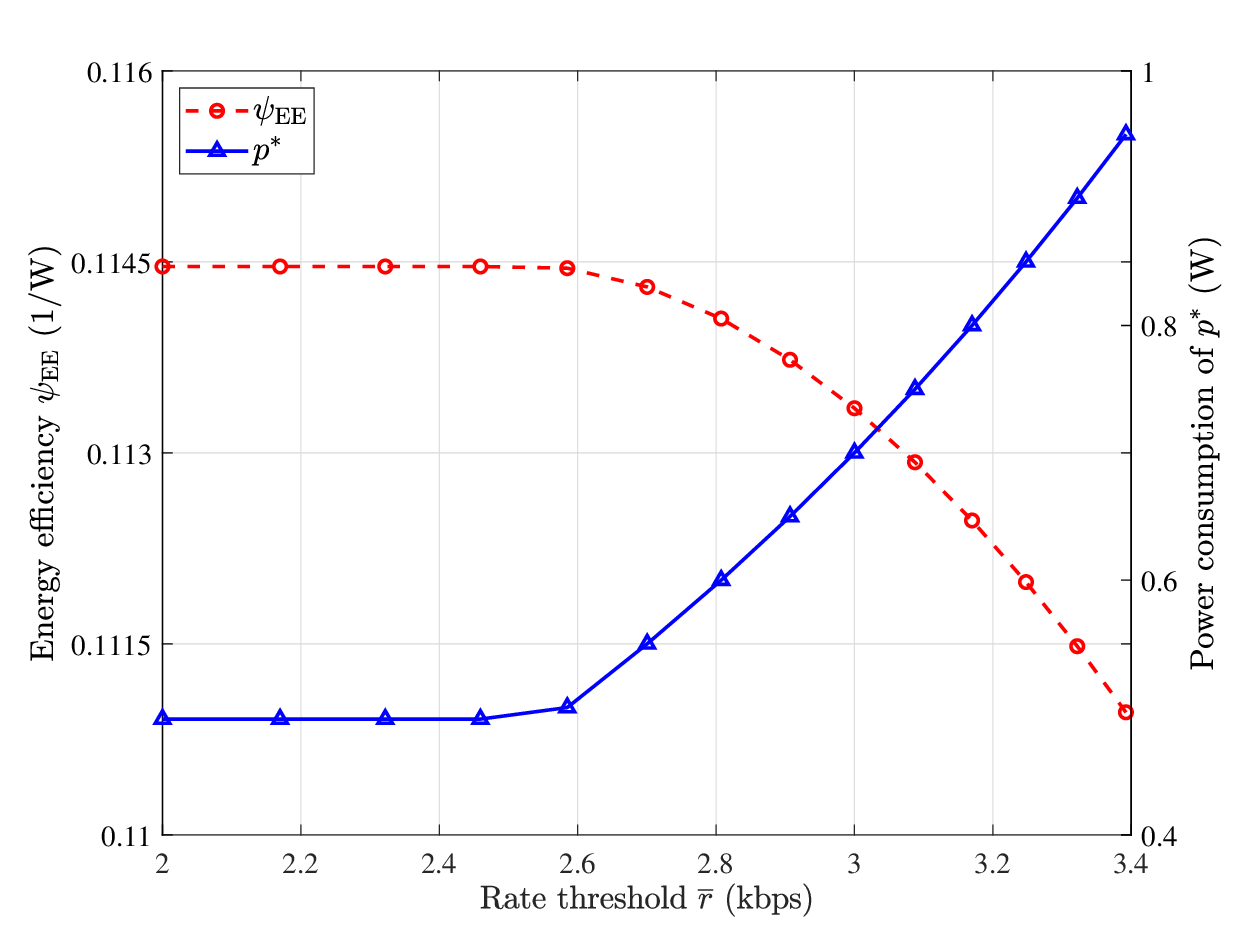} 
		\vskip-0.1cm
		\centering {(b)}
	\end{minipage}
	\captionsetup{justification=justified}
	\caption{  { (a) Power consumption  $p^*$ and energy efficiency ${\psi_{\rm{EE}}}$ versus power threshold $P_{\rm{U}}$; (b) Power consumption  $p^*$ and energy efficiency $\psi_{\rm{EE}}$ versus rate threshold $\overline r$. } }
	\label{DT}
\end{figure}

 \subsection{Evaluation of Energy Efficiency}

In the following, we present simulation results to evaluate the optimal power control for maximizing energy efficiency in semantic communication systems.

 \begin{figure*}[!t]
	\centering
	\begin{minipage}[t]{0.3\textwidth}
		\centering
		\includegraphics[width=\textwidth]{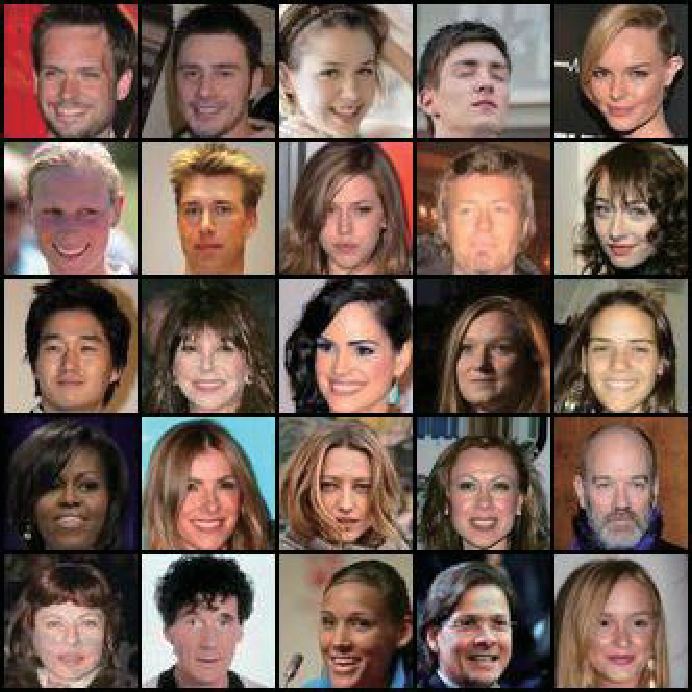}
		\vskip-0.1cm\centering {\footnotesize (a)}
	\end{minipage}
	\begin{minipage}[t]{0.3\textwidth}
		\centering
		\includegraphics[width=\textwidth]{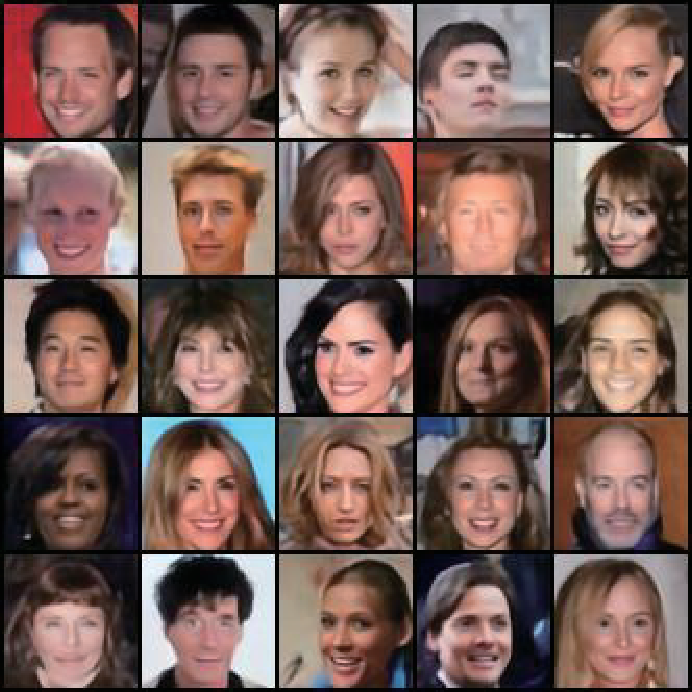}
		\vskip-0.1cm\centering {\footnotesize (b) }
	\end{minipage}
	\begin{minipage}[t]{0.3\textwidth}
		\centering
		\includegraphics[width=\textwidth]{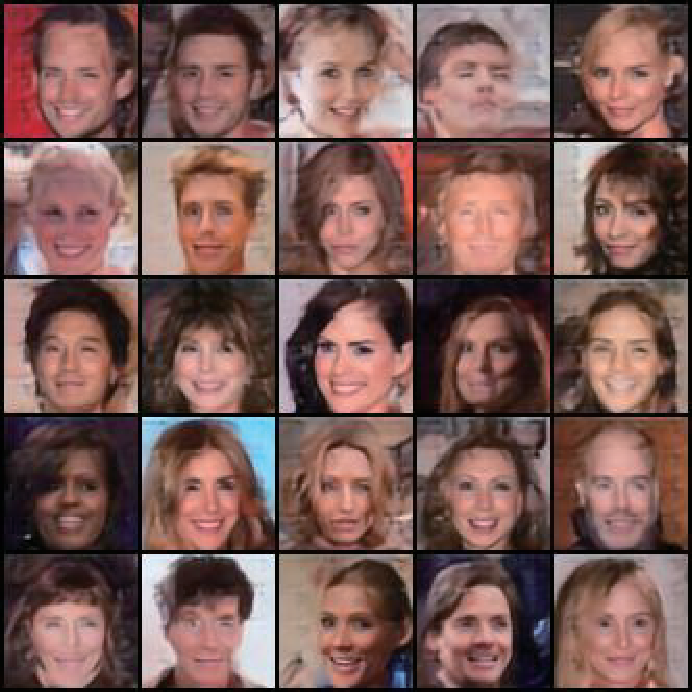}
		\vskip-0.1cm\centering {\footnotesize (c) }
	\end{minipage}
	\captionsetup{justification=justified}
	\caption{ The    worst image reconstruction performance of  different power allocation schemes. (a) Original images; (b) Proposed power allocation scheme; (c) Fixed power allocation scheme; }
	\label{bis_vs_aver}
\end{figure*}

 \begin{figure}
	 	\centering
	 	\includegraphics[width=0.4\textwidth]{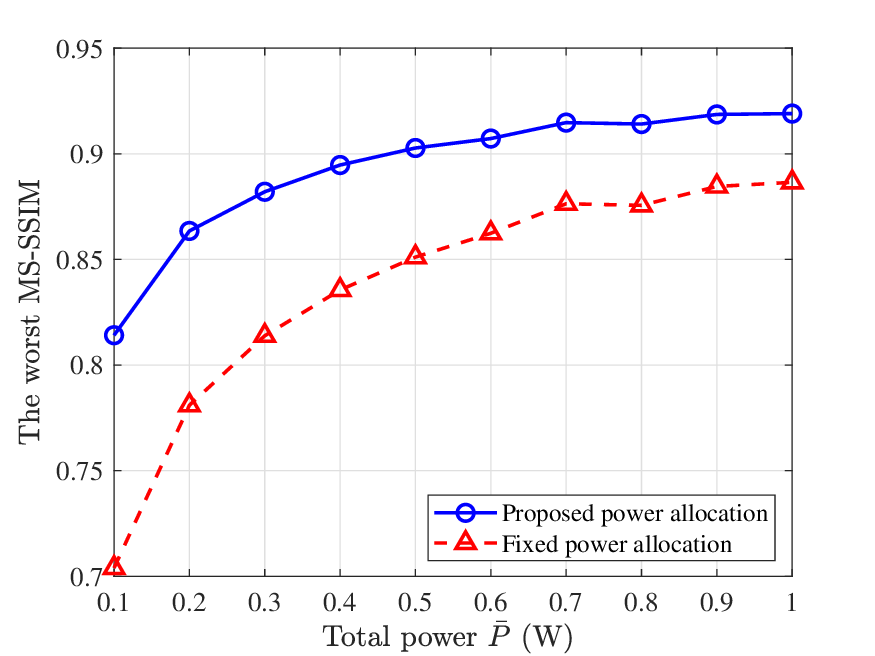}
		\captionsetup{justification=justified}
	 	\caption{The performance comparison between proposed and fixed power allocation showing the worst MS-SSIM in multi-user OFDMA downlink semantic communication.}
	 	\label{Min_MS_SSIM}
\end{figure}

Fig.\ref{DT} (a) illustrates the energy efficiency ${\psi _{{\rm{EE}}}}$ and power consumption  $p^*$ versus the power threshold ${P_{\rm{U}}}$ with a fixed rate threshold $\overline{r} {= 2 \rm{kbps}}$. As ${P_{\rm{U}}}$ increases, the power consumption  $p^*$ first increases and then remains constant. This is because when the power threshold ${P_{\rm{U}}}$ is lower than ${\tilde{p}}$, $p^*$ is limited by ${P_{\rm{U}}}$, whereas when ${P_{\rm{U}}}$ is higher than ${\tilde{p}}$, $p^*$ reaches ${\tilde{p}}$, and the power threshold constraint ${P_{\rm{U}}}$ becomes inactive. Moreover, the energy efficiency ${\psi _{{\rm{EE}}}}$ increases as ${P_{\rm{U}}}$ increases, reaching its optimal value and then remaining constant when ${P_{\rm{U}}}$ is higher than ${\tilde{p}}$.

\begin{table}[!t]
	\captionsetup{justification=justified}
	\caption{The worst reconstructed quality of the proposed power allocation scheme and the fixed power allocation scheme with the same inputs.}
	\centering
	\small
	\scalebox{1}{
		\resizebox{0.9\linewidth}{!}{\begin{tabular}{|c|c|c|}
				\hline
				\rule{0pt}{15pt}	 & Fixed power allocation & Proposed power allocation  \\ \hline
				\rule{0pt}{15pt}  	PSNR(dB) & 20.45 & 24.26    \\ \hline
				\rule{0pt}{15pt}  	MS-SSIM & 0.818 & 0.924  \\ \hline
		\end{tabular}}
		\label{table10}
	}
\end{table}

Fig.\ref{DT} (b) depicts the energy efficiency ${\psi _{{\rm{EE}}}}$ and power consumption  $p^*$ versus the minimum rate threshold versus ${\tilde{p}} { = 10 \rm{W}}$ with the fixed power threshold ${P_{\rm{U}}}$. As shown in Fig. \ref{DT}(b), the power consumption  $p^*$ first remains constant then increases as the rate threshold $\overline{r}$ increases. 
This is because when the rate $\overline{r}$ is lower than approximately 2.46 kbps, the optimal power consumption $p^*$ remains fixed at $\tilde{p}$. However, when the rate $\overline{r}$ exceeds 2.46 kbps, the optimal power consumption $p^*$ increases, requiring more power to support the rate constraint. Additionally, as the rate threshold $\overline{r}$ increases, the energy efficiency $\psi_{\text{EE}}$ first remains constant and then decreases. This is because $p^*$ is equal to $\tilde{p}$ when the rate $\overline{r}$ is less than or equal to 2.46 kbps, while $p^*$ is greater than $\tilde{p}$ when the rate $\overline{r}$ is greater than 2.46 kbps.


 \subsection{Evaluation of Power Allocation for Multi-user Case}

Fig. \ref{bis_vs_aver} compares the semantic decoded images performance of the proposed power allocation scheme and the fixed power allocation scheme for the  multi-user OFDMA downlink semantic communication network with three users, i.e., $K=3$.
Specifically,  Fig. \ref{bis_vs_aver} (a) shows the original images, and Fig. \ref{bis_vs_aver} (b) and  (c) show  the worst semantic decoded images of the  multiple users of  the proposed power allocation scheme and the fixed power allocation scheme, respectively.
Comparing  Fig. \ref{bis_vs_aver} (b) and  (c), it can be seen that the semantic decoding image quality of our proposed power allocation scheme is significantly higher than that of the fixed power allocation scheme.
Moreover, as listed in Table \ref{table10}, the MS-SSIM and PSNR  of reconstructed images of  the fixed   power allocation scheme  are $0.818$ and $20.45$, while the MS-SSIM and PSNR  of reconstructed images of  the proposed power allocation scheme are $0.924$ and $24.26$, respectively.
The results in Fig. \ref{bis_vs_aver}  and Table \ref{table10}  show the effectiveness and superiority of the proposed    power allocation schemes.
                                            
Fig. \ref{Min_MS_SSIM} compares  the image reconstruction  performance of   the proposed power allocation scheme and the fixed power allocation scheme versus total transmitted power $\overline P$ for the  multi-user OFDMA downlink semantic
communication network.
Fig. \ref{Min_MS_SSIM} shows that the minimum MS-SSIM of the proposed power allocation scheme is clearly higher than that of  the   fixed power allocation scheme for all SNR, especially at low total transmitted power $\overline P$, where the advantage of the proposed power allocation scheme is more prominent. Fig. \ref{Min_MS_SSIM}  verifies the effectiveness and superiority of the proposed    power allocation schemes.

\section{Conclusion}
In this work,  we proposed an  ABG formula to   model the  relationship between the performance metric and SNR for  semantic communications with image reconstruction tasks and inference tasks.
Moreover, we proposed a  empirical formula with closed-form expression to fit this relationship.
To  our best knowledge, this ABG formula is the first  empirical expression between  end-to-end performance metrics and   SNR  for  semantic communications.
Particularly,  the proposed ABG  formula  can well fit the commonly used DL based semantic communications  with image   reconstruction tasks and inference tasks.
Furthermore, we designed  an adaptive power control scheme  for semantic communications over random fading channels, which can effectively guarantee the MS-SSIM requirements. Then, we developed an optimal power allocation scheme to maximize the  energy efficiency of the semantic communication system.
In addition, we developed the power allocation scheme to maximize the minimum  MS-SSIM  of multiple users for OFDMA downlink semantic communication networks.		 						 
Finally, we demonstrated both the effectiveness and superiority of the proposed ABG formula and  power allocation schemes.

%
%
%
\bibliographystyle{IEEEtran}

\bibliography{ABG_arXiv}

\begin{thebibliography}{10}
\providecommand{\url}[1]{#1}
\csname url@samestyle\endcsname
\providecommand{\newblock}{\relax}
\providecommand{\bibinfo}[2]{#2}
\providecommand{\BIBentrySTDinterwordspacing}{\spaceskip=0pt\relax}
\providecommand{\BIBentryALTinterwordstretchfactor}{4}
\providecommand{\BIBentryALTinterwordspacing}{\spaceskip=\fontdimen2\font plus
\BIBentryALTinterwordstretchfactor\fontdimen3\font minus
  \fontdimen4\font\relax}
\providecommand{\BIBforeignlanguage}[2]{{%
\expandafter\ifx\csname l@#1\endcsname\relax
\typeout{** WARNING: IEEEtran.bst: No hyphenation pattern has been}%
\typeout{** loaded for the language `#1'. Using the pattern for}%
\typeout{** the default language instead.}%
\else
\language=\csname l@#1\endcsname
\fi
#2}}
\providecommand{\BIBdecl}{\relax}
\BIBdecl

\bibitem{Mihai_CST_2022}
S.~Mihai, M.~Yaqoob, D.~V. Hung, W.~Davis, P.~Towakel, M.~Raza, M.~Karamanoglu,
  B.~Barn, D.~Shetve, R.~V. Prasad, H.~Venkataraman, R.~Trestian, and H.~X.
  Nguyen, ``Digital twins: A survey on enabling technologies, challenges,
  trends and future prospects,'' \emph{IEEE Commun. Surv. Tutor.}, vol.~24,
  no.~4, pp. 2255--2291, Sept. 2022.

\bibitem{Lee_2024}
L.-H. Lee, T.~Braud, P.~Y. Zhou, L.~Wang, D.~Xu, Z.~Lin, A.~Kumar, C.~Bermejo,
  and P.~Hui, \emph{All One Needs to Know about Metaverse: A Complete Survey on
  Technological Singularity, Virtual Ecosystem, and Research Agenda}, 2024.

\bibitem{Dai_WC_2023}
J.~Dai, P.~Zhang, K.~Niu, S.~Wang, Z.~Si, and X.~Qin, ``Communication beyond
  transmitting bits: Semantics-guided source and channel coding,'' \emph{IEEE
  Wireless Commun.}, vol.~30, no.~4, pp. 170--177, Aug. 2023.

\bibitem{Strinati_CN_2021}
\BIBentryALTinterwordspacing
E.~{Calvanese Strinati} and S.~Barbarossa, ``6{G} networks: Beyond shannon
  towards semantic and goal-oriented communications,'' \emph{Comput. Networks},
  vol. 190, p. 107930, Mar. 2021. [Online]. Available:
  \url{https://www.sciencedirect.com/science/article/pii/S1389128621000773}
\BIBentrySTDinterwordspacing

\bibitem{Lu_WC_2024}
K.~Lu, Q.~Zhou, R.~Li, Z.~Zhao, X.~Chen, J.~Wu, and H.~Zhang, ``Rethinking
  modern communication from semantic coding to semantic communication,''
  \emph{IEEE Wireless Commun.}, vol.~30, no.~1, pp. 158--164, May. 2023.

\bibitem{Patel_WE_2022}
A.~Patel, N.~C. Debnath, and P.~K. Shukla, ``Secureont: A security ontology for
  establishing data provenance in semantic web,'' \emph{Journal of Web
  Engineering}, vol.~21, no.~4, pp. 1347--1370, Jun. 2022.

\bibitem{Lo_WCL_2023}
W.~F. Lo, N.~Mital, H.~Wu, and D.~Gündüz, ``Collaborative semantic
  communication for edge inference,'' \emph{IEEE Wireless Commun. Lett.},
  vol.~12, no.~7, pp. 1125--1129, Mar. 2023.

\bibitem{Nam_ICASSP_2024}
H.~Nam, J.~Park, J.~Choi, M.~Bennis, and S.-L. Kim, ``Language-oriented
  communication with semantic coding and knowledge distillation for
  text-to-image generation,'' in \emph{IEEE International Conference on
  Acoustics, Speech and Signal Processing (ICASSP)}, Mar. 2024, pp.
  13\,506--13\,510.

\bibitem{Qiao_WCL_2024}
L.~Qiao, M.~B. Mashhadi, Z.~Gao, C.~H. Foh, P.~Xiao, and M.~Bennis,
  ``Latency-aware generative semantic communications with pre-trained diffusion
  models,'' \emph{IEEE Wireless Commun. Lett.}, vol.~13, no.~10, pp.
  2652--2656, Oct. 2024.

\bibitem{Yang_ICASSP_2023}
K.~Yang, S.~Wang, J.~Dai, K.~Tan, K.~Niu, and P.~Zhang, ``Witt: A wireless
  image transmission transformer for semantic communications,'' in \emph{ICASSP
  2023 - 2023 IEEE International Conference on Acoustics, Speech and Signal
  Processing (ICASSP)}, May. 2023, pp. 1--5.

\bibitem{Ma_TWC_2023}
S.~Ma, W.~Qiao, Y.~Wu, H.~Li, G.~Shi, D.~Gao, Y.~Shi, S.~Li, and N.~Al-Dhahir,
  ``Task-oriented explainable semantic communications,'' \emph{IEEE Trans.
  Wireless Commun.}, vol.~22, no.~12, pp. 9248--9262, Apr. 2023.

\bibitem{Xie_TSP_2021}
H.~Xie, Z.~Qin, L.~Geoffrey, Ye, and B.-H. Juang, ``Deep learning enabled
  semantic communication systems,'' \emph{IEEE Trans. Signal Process.},
  vol.~69, pp. 2663--2675, Apr. 2021.

\bibitem{Farsad_ACASSP_2018}
N.~Farsad, M.~Rao, and A.~Goldsmith, ``Deep learning for joint source-channel
  coding of text,'' in \emph{Proc. IEEE Int. Conf. Acoust., Speech Signal
  Process. (ICASSP)}, Sept. 2018, pp. 2326--2330.

\bibitem{Xie_JSAC_2020}
H.~Xie and Z.~Qin, ``A lite distributed semantic communication system for
  internet of things,'' \emph{IEEE J. Sel. Areas Commun.}, vol.~39, no.~1, pp.
  142--153, Jan. 2020.

\bibitem{Yi_TWC_2023}
P.~Yi, Y.~Cao, X.~Kang, and Y.-C. Liang, ``Deep learning-empowered semantic
  communication systems with a shared knowledge base,'' \emph{IEEE Trans.
  Wireless Commun.}, pp. 1--1, Nov. 2023.

\bibitem{Jia_CL_2023}
Y.~Jia, Z.~Huang, K.~Luo, and W.~Wen, ``Lightweight joint source-channel coding
  for semantic communications,'' \emph{IEEE Commun. Lett.}, vol.~27, no.~12,
  pp. 3161--3165, Nov. 2023.

\bibitem{Weng_JSAC_2021}
Z.~Weng and Z.~Qin, ``Semantic communication systems for speech transmission,''
  \emph{IEEE J. Sel. Areas Commun}, vol.~39, no.~8, pp. 2434--2444, Jun. 2021.

\bibitem{Han_JSAC_2023}
T.~Han, Q.~Yang, Z.~Shi, S.~He, and Z.~Zhang, ``Semantic-preserved
  communication system for highly efficient speech transmission,'' \emph{IEEE
  J. Sel. Areas Commun.}, vol.~41, no.~1, pp. 245--259, Jan. 2023.

\bibitem{Huang_JSAC_2023}
D.~Huang, F.~Gao, X.~Tao, Q.~Du, and J.~Lu, ``Toward semantic communications:
  Deep learning-based image semantic coding,'' \emph{IEEE J. Sel. Areas
  Commun.}, vol.~41, no.~1, pp. 55--71, Nov. 2023.

\bibitem{Zhang_JSAC_2023}
H.~Zhang, S.~Shao, M.~Tao, X.~Bi, and K.~B. Letaief, ``Deep learning-enabled
  semantic communication systems with task-unaware transmitter and dynamic
  data,'' \emph{IEEE J. Sel. Areas Commun.}, vol.~41, no.~1, pp. 170--185, Nov.
  2023.

\bibitem{Erdemir_JSAC_2023}
E.~Erdemir, T.-Y. Tung, P.~L. Dragotti, and D.~Gündüz, ``Generative joint
  source-channel coding for semantic image transmission,'' \emph{IEEE J. Sel.
  Areas Commun.}, vol.~41, no.~8, pp. 2645--2657, Jun. 2023.

\bibitem{Huang_IoT_2024}
J.~Huang, D.~Li, C.~Huang, X.~Qin, and W.~Zhang, ``Joint task and data-oriented
  semantic communications: A deep separate source-channel coding scheme,''
  \emph{IEEE Internet Things J.}, vol.~11, no.~2, pp. 2255--2272, Jan. 2024.

\bibitem{Kang_JSAC_2023}
J.~Kang, H.~Du, Z.~Li, Z.~Xiong, S.~Ma, D.~Niyato, and Y.~Li, ``Personalized
  saliency in task-oriented semantic communications: Image transmission and
  performance analysis,'' \emph{IEEE J. Sel. Areas Commun.}, vol.~41, no.~1,
  pp. 186--201, Nov. 2023.

\bibitem{Tung_JSAC_2022}
T.-Y. Tung and D.~Gündüz, ``Deepwive: Deep-learning-aided wireless video
  transmission,'' \emph{IEEE J. Sel. Areas Commun.}, vol.~40, no.~9, pp.
  2570--2583, Jul. 2022.

\bibitem{Jiang_JSAC_2023}
P.~Jiang, C.-K. Wen, S.~Jin, and G.~Y. Li, ``Wireless semantic communications
  for video conferencing,'' \emph{IEEE J. Sel. Areas Commun.}, vol.~41, no.~1,
  pp. 230--244, Jan. 2023.

\bibitem{Wang_JSAC_2023}
S.~Wang, J.~Dai, Z.~Liang, K.~Niu, Z.~Si, C.~Dong, X.~Qin, and P.~Zhang,
  ``Wireless deep video semantic transmission,'' \emph{IEEE J. Sel. Areas
  Commun.}, vol.~41, no.~1, pp. 214--229, Jan. 2023.

\bibitem{Dai_JSAC_2022}
J.~Dai, S.~Wang, K.~Tan, Z.~Si, X.~Qin, K.~Niu, and P.~Zhang, ``Nonlinear
  transform source-channel coding for semantic communications,'' \emph{IEEE J.
  Sel. Areas Commun.}, vol.~40, no.~8, pp. 2300--2316, Aug. 2022.

\bibitem{lee_Access_2019}
C.-H. Lee, J.-W. Lin, P.-H. Chen, and Y.-C. Chang, ``Deep learning-constructed
  joint transmission-recognition for internet of things,'' \emph{IEEE Access},
  vol.~7, pp. 76\,547--76\,561, Jun. 2019.

\bibitem{Huang_JSAC_2022}
D.~Huang, F.~Gao, X.~Tao, Q.~Du, and J.~Lu, ``Toward semantic communications:
  Deep learning-based image semantic coding,'' \emph{IEEE J. Sel. Areas
  Commun.}, vol.~41, no.~1, pp. 55--71, Nov. 2022.

\bibitem{Hu_TWC_2023}
Q.~Hu, G.~Zhang, Z.~Qin, Y.~Cai, G.~Yu, and G.~Y. Li, ``Robust semantic
  communications with masked vq-vae enabled codebook,'' \emph{IEEE Trans.
  Wireless Commun.}, vol.~22, no.~12, pp. 8707--8722, Apr. 2023.

\bibitem{Shao_JSAC_2022}
Y.~M. J.~Shao and J.~Zhang, ``Learning task-oriented communication for edge
  inference: An information bottleneck approach,'' \emph{IEEE J. Sel. Areas.
  Commun.}, vol.~40, no.~1, pp. 197--211, Jan. 2022.

\bibitem{Xie_JSAC_2023}
S.~Xie, S.~Ma, M.~Ding, Y.~Shi, M.~Tang, and Y.~Wu, ``Robust information
  bottleneck for task-oriented communication with digital modulation,''
  \emph{IEEE J. Sel. Areas Commun.}, vol.~41, no.~8, pp. 2577--2591, Aug. 2023.

\bibitem{Tishby_arXiv_2000}
N.~Tishby, F.~C. Pereira, and W.~Bialek, ``The information bottleneck method,''
  \emph{arXiv preprint physics/0004057}, 2000.

\bibitem{Hu_2022}
Q.~Hu, G.~Zhang, Z.~Qin, Y.~Cai, G.~Yu, and G.~Y. Li, ``Robust semantic
  communications against semantic noise,'' pp. 1--6, Jan. 2022.

\bibitem{Zhang_TWC_2023}
W.~Zhang, H.~Zhang, H.~Ma, H.~Shao, N.~Wang, and V.~C.~M. Leung, ``Predictive
  and adaptive deep coding for wireless image transmission in semantic
  communication,'' \emph{IEEE Trans. Wireless Commun.}, vol.~22, no.~8, pp.
  5486--5501, Jan. 2023.

\bibitem{Jiang_TC_2022}
P.~Jiang, C.~K. Wen, S.~Jin, and G.~Y. Li, ``Deep source-channel coding for
  sentence semantic transmission with {HARQ},'' \emph{IEEE Trans. Commun.},
  vol.~70, no.~8, pp. 5225--5240, Jun. 2022.

\bibitem{Peng_GCC_2022}
X.~Peng, Z.~Qin, D.~Huang, X.~Tao, J.~Lu, G.~Liu, and C.~Pan, ``A robust deep
  learning enabled semantic communication system for text,'' in \emph{GLOBECOM
  2022 - 2022 IEEE Global Communications Conference}, Jan. 2022, pp.
  2704--2709.

\bibitem{Xiao_ICASSP_2023}
Z.~Xiao, S.~Yao, J.~Dai, S.~Wang, K.~Niu, and P.~Zhang, ``Wireless deep speech
  semantic transmission,'' in \emph{ICASSP 2023 - 2023 IEEE International
  Conference on Acoustics, Speech and Signal Processing (ICASSP)}, May. 2023,
  pp. 1--5.

\bibitem{Yang_JSTSP_2021}
R.~Yang, F.~Mentzer, L.~Van~Gool, and R.~Timofte, ``Learning for video
  compression with recurrent auto-encoder and recurrent probability model,''
  \emph{IEEE J. Sel. Top. Signal Process.}, vol.~15, no.~2, pp. 388--401, Dec.
  2021.

\bibitem{Xie_2023}
S.~Xie, S.~Ma, M.~Ding, Y.~Shi, M.~Tang, and Y.~Wu, ``Robust information
  bottleneck for task-oriented communication with digital modulation,''
  \emph{IEEE J. Sel. Areas Commun.}, vol.~41, no.~8, pp. 2577--2591, Aug. 2023.

\bibitem{Zhang_2023_Practical}
K.~Zhang, Y.~Li, J.~Liang, J.~Cao, Y.~Zhang, H.~Tang, D.-P. Fan, R.~Timofte,
  and L.~V. Gool, ``Practical blind image denoising via swin-conv-unet and data
  synthesis,'' \emph{Mach. Intell. Res.}, vol.~20, no.~6, Sept.

\bibitem{Dosovitskiy_2020_arXiV}
A.~Dosovitskiy, ``An image is worth 16x16 words: Transformers for image
  recognition at scale,'' \emph{arXiv preprint arXiv:2010.11929}, 2020.

\bibitem{Yang_TCCN_2024}
K.~Yang, S.~Wang, J.~Dai, X.~Qin, K.~Niu, and P.~Zhang, ``Swinjscc: Taming swin
  transformer for deep joint source-channel coding,'' \emph{IEEE Trans. Cognit.
  Commun. Networking}, pp. 1--1, 2024.

\bibitem{Thinsungnoena_L_2015}
T.~Thinsungnoena, N.~Kaoungkub, P.~Durongdumronchaib, K.~Kerdprasopb,
  N.~Kerdprasopb \emph{et~al.}, ``The clustering validity with silhouette and
  sum of squared errors,'' \emph{learning}, vol.~3, no.~7, Jan. 2015.

\bibitem{Zappone_2015}
A.~Zappone and E.~Jorswieck, ``Energy efficiency in wireless networks via
  fractional programming theory,'' \emph{Foundations \& Trends in
  Communications \& Information Theory}, vol.~11, no. 3-4, pp. 185--396, Jun.
  2015.

\bibitem{Matlab_2012}
S.~Matlab, ``Matlab,'' \emph{The MathWorks, Natick, MA}, 2012.

\bibitem{Litchfield_JPET_1949}
J.~j. Litchfield and F.~Wilcoxon, ``A simplified method of evaluating
  dose-effect experiments,'' \emph{J. Pharmacol. Exp. Ther.}, vol.~96, no.~2,
  Jun.

\end{thebibliography}

\end{document}